# *Continental-scale streamflow modeling of basins with reservoirs: towards a coherent deep-learning-based strategy*

Wenyu Ouyang[1], Kathryn Lawson[2], Dapeng Feng[2], Lei Ye[1], Chi Zhang[1], Chaopeng Shen[2,*]

[1] School of Hydraulic Engineering, Dalian University of Technology, Dalian, China

[2] Civil and Environmental Engineering, Pennsylvania State University, University Park, PA, USA

## Abstract

A large fraction of major waterways have dams influencing streamflow, which must be accounted for in large-scale hydrologic modeling. However, daily streamflow prediction for basins with dams is challenging for various modeling approaches, especially at large scales. Here we examined which types of dammed basins could be well represented by long short-term memory (LSTM) models using readily-available information, and delineated the remaining challenges. We analyzed data from 3557 basins (83% dammed) over the contiguous United States and noted strong impacts of reservoir purposes, degree of regulation (*dor*), and diversion on streamflow modeling. While a model trained on a widely-used reference-basin dataset performed poorly for non-reference basins, the model trained on the whole dataset presented a median Nash-Sutcliffe efficiency coefficient (NSE) of 0.74. The zero-*dor*, small-*dor* (with storage of approximately a month of average streamflow or less), and large-*dor* basins were found to have distinct behaviors, so migrating models between categories yielded catastrophic results, which means we must not treat small-*dor* basins as reference ones. However, training with pooled data from different sets yielded optimal median NSEs of 0.72, 0.79, and 0.64 for these respective groups, noticeably stronger than existing models. These results support a coherent modeling strategy where smaller dams (storing about a month of average streamflow or less) are modeled implicitly as part of basin rainfall-runoff processes; then, large-*dor* reservoirs of certain types can be represented explicitly. However, dammed basins must be present in the training dataset. Future work should examine separate modeling of large reservoirs for fire protection and irrigation, hydroelectric power generation, and flood control.

---

* corresponding author. Chaopeng Shen: cshen@engr.psu.edu

# 1. Introduction.

Two-thirds of the longest rivers in the world are not flowing freely (Grill et al., 2019): more than 800,000 dammed reservoirs impede the world's rivers, including 90,000 in the United States (International Rivers, 2007). Dams exert significant control on streamflows by changing the magnitude and timing of the discharges (Gutenson et al., 2020). The ability to anticipate upstream reservoir operations at a daily scale has significant operational value for optimal water resources management.

For large-scale hydrologic modeling at the daily scale, we need accurate and tractable methods to account for the influence of small and large reservoirs on streamflow. One may use a *reservoir-centric* modeling approach, in which each reservoir needs to be represented explicitly with its own characteristics, operational rules, storage, inflow, and outflow. This approach may not scale well to large scales, however, as there may be dozens or even hundreds of reservoirs upstream of the outlet of a large basin. A different approach would be *basin-centric* (or grid-centric, also called lumped), in which all the reservoirs in a subbasin (or a computational gridcell) are grouped together into one unit in the river routing module. Apparently, the basin-centric (or lumped) paradigm can vastly reduce modeling complexity (Ehsani et al., 2016; Payan et al., 2008). Alternatively, a mixed approach can be taken where some reservoirs are lumped while some others are explicitly represented. Current large-scale hydrologic models such as the National Water Model (NWM) (Gochis et al., 2018), or land surface hydrologic models with routing schemes, e.g. the Community Land Model (Lawrence et al., 2019) simulate some major reservoirs and make the habitual assumption of ignoring the smaller reservoirs. The questions are then: (i) What kinds of reservoirs can be modeled in a lumped fashion and what kind cannot? (ii) Can we ignore the impacts of small reservoirs and assume they are behaviorally similar to undammed basins?

It has been difficult to reliably obtain strong model performance for dammed basins using a rule-based system at large scales. From a literature survey (see more details in Appendix Table S1), it seems difficult to obtain Nash-Sutcliffe model efficiency coefficient (NSE) values that are higher than 0.65 by assuming generic reservoir operational schemes (Biemans et al., 2011; Hanasaki et al., 2006; Shin et al., 2019; Voisin et al., 2013). Hanasaki et al. (2006) derived a demand-driven approach for global reservoir routing and laid the foundation for subsequent developments, showing error reduction compared to no-reservoir simulations, but no NSE was reported. Voison et al. (2013) improved upon the formulation from Hanasaki et al. (2006) to the heavily dammed Columbia River Basin and reported decent correlation but mostly negative NSEs, indicating substantial biases. Unlike generic release schemes, empirically derived target storage-release functions can be parameterized for individual reservoirs with sufficiently long observational records of releases, inflows, and storage levels, and can reproduce observed flows more accurately (Kim et al., 2020; Turner et al., 2020; Wu and Chen, 2012; Yassin et al., 2019; Zajac et al., 2017; Zhao et al., 2016). Yassin et al. (2019) used piecewise-linear relationships between reservoir storage, inflow, and release to describe reservoir policies and obtained a median NSE of ~0.5 for 37 reservoirs across the globe. Zajac et al. (2017) reported a maximum NSE of 0.61 for 390 stations around the world. Although these results represent significant progress in research, further research was still needed to inform whether these improvements were robust when simulated inflows from the hydrologic models, rather than observed inflows, were used as the input to reservoir modules at large scales (Turner et al., 2020). In addition, one can certainly argue the current performance levels left room for improvement, which can provide better utility for practical applications.

Artificial neural networks (ANNs) and other machine learning models have been applied to establish data-driven rules that relate reservoir storage, inflow, and release data. Ehsani et al. (2016) used ANNs to predict daily release using previous days' reservoir storage volume along with inflow and release measurements, and reported an NSE of 0.86. Yang et al. (2019) similarly applied recurrent neural networks, using inflow and water storage as inputs, to simulate the daily operation of three multi-purpose reservoirs located in one basin, and reported an NSE value over 0.85. However, the use of recent storage and inflow data is akin to a form of data assimilation and is known to greatly improve simulations for short-term forecast (Feng et al., 2020a), but we do not use recent observations here as our objective is long-term projection. In addition, the existing generally-available reservoir databases (Lehner et al., 2011; Mulligan et al., 2020; Patterson and Doyle, 2018) mainly provide information on dam design specifications or operational details for some of the most significant reservoirs, which is not available for large-scale modeling in dammed basins.

Recently, the long short-term memory (LSTM) network (Hochreiter and Schmidhuber, 1997), a deep learning (DL) algorithm, has been applied to explore the ability to predict streamflow in basins across the CONUS. It is relatively inexpensive (in terms of time) to apply at large spatial scales, and has grown to be a well-established hydrologic modeling tool (Shen, 2018). LSTM-based models can effectively learn streamflow dynamics, and have shown superior performance compared to other hydrological benchmark models (Ayzel et al., 2020; Feng et al., 2020a; Kratzert et al., 2019b). For example, Kratzert et al. (2019b) reported that the median NSE value in the evaluation period could reach 0.74 for a 531-basin subset of the 671-basin Catchment Attributes and Meteorology for Large-Sample Studies (CAMELS) dataset using the forcing data from North American Land Data Assimilation (NLDAS) system. More recently, Feng et al. (2020a) improved the forecast NSE median to 0.86 with the addition of a data integration kernel which incorporated recent discharge observations. However, the CAMELS dataset, which all these studies were based on, is composed of basins that are considered to be "reference" or undisturbed basins, which have minimal anthropogenic impacts (i.e., minimal land use changes, minimal human water withdrawals) (Addor et al., 2017; Newman et al., 2015). To our knowledge, there is no systematic knowledge regarding how LSTM performs in basins with significant human modifications such as reservoirs or water diversion, especially at large scales.

Here we followed a divide-and-conquer approach to tackle the difficult problem of long-term daily streamflow prediction from dammed basins, and to delineate where challenges reside. We addressed the following questions: (1) Given only generally-available reservoir information, how well can LSTM networks make long-term daily streamflow predictions for basins with reservoirs across the entire CONUS? (2) How differently do basins with or without reservoirs of different sizes function in streamflow --- how much error are we making if we simply ignore small reservoirs and treat those basins with small reservoirs as reference basins? (3) What kinds of reservoirs (purpose, size, diversion) can be well modeled in a lumped fashion and what kinds cannot? These questions have not been answered in the literature and the answers will help the community to devise an informed and coherent modeling strategy. We further provide experiences to the community on how to best form an appropriate training dataset, e.g., whether we should include basins with or without reservoirs and whether we should stratify basins into different categories based on reservoir characteristics, or simply group them together.

## 2. Methods

As an overview, LSTM-based models were trained to predict long-term daily streamflow from basins with or without reservoirs. The inputs include atmospheric forcing time series data and static basin attributes (physiographic attributes and anthropogenic influences). We trained the models using various subsets from a newly compiled 3557-basin dataset across the CONUS as well as the CAMELS dataset. Basins with complete streamflow records from 1 January 1990 through 31 December 2009 were selected from the Geospatial Attributes of Gages for Evaluating Streamflow II (GAGES-II) dataset (Falcone, 2011). Below we provide the details of the procedures.

### 2.1. LSTM

Long Short-Term Memory (LSTM) networks are a special kind of recurrent neural network (RNN) which can both learn from sequential data and address the notorious exploding and/or vanishing gradient problem (Hochreiter, 1998). These networks are composed of memory cells, the keys to which are the "cell states" and "gates" that control information flow within the LSTM algorithm. Cell states allow information to be stored over long time periods, which is important for modeling catchment processes like snow, subsurface flow, and reservoir storage. Based on the input of the current time step and the output from the previous one, a "forget gate" decides what information is going to be removed from the existing cell state. Next, a sigmoid layer and a tanh layer are applied as an "input gate" to update the cell state. Finally, the cell state is put through a tanh function and multiplied by the output of the sigmoid "output gate" to determine the final output.

There are different formulations of LSTM-based models. Kratzert et al. (2019b) used an N-to-1 model to predict streamflow, which means that the input was a multi-step time series and the output was a one-step variable. An N-to-M LSTM-based model, also called a sequence-to-sequence model, was employed to predict multi-time-step streamflows by Xiang et al. (2020). In the present study, following Feng et al. (2020a), we trained a CONUS-scale N-to-N model using meteorological forcings and static attributes of the basins to predict daily discharge. Here we did not use discharge from previous days as inputs. We trained the model on sequences of a fixed length (365 days), but for inference, we ran the model in a single forward pass through the full time period. This procedure means that during training, the LSTM has no context for the initial input steps of each sequence. However, in our preliminary anaysis, we added a warm-up period but found it to not have any noticeable impact. Thus we neglected the warm-up period for performance reasons. The N-to-N model had significant advantages in efficiency, and could reach convergence for the 671 basins in the CAMELS dataset with 10 years of training data in 69 minutes on an NVIDIA 1080 Ti graphical processing unit (GPU). In this paper, the model was able to be trained on 10-year data for the entire 3557-basin dataset until convergence was achieved (300 epochs) in 427 minutes of computational time. In our code, we randomly sampled for sites and training periods to form mini-batches and we defined the total number of iterations in an epoch as corresponding to the probability that 99% of the time periods of all basins are picked in the epoch.

The forward propagation equations of the present LSTM-based model can be summarized as the following (see Figure S1 in Appendix for more details), based on the notations in Fang et al. (2020).

$$x^{(t)} = ReLU\left(W_{xx}x_0^{(t)} + b_{xx}\right) \quad (1)$$

$$f^{(t)} = \sigma\left(D(W_{fx}x^{(t)}) + D(W_{fh}h^{(t-1)}) + b_f\right) \quad (2)$$

$$i^{(t)} = \sigma\left(D(W_{ix}x^{(t)}) + D(W_{ih}h^{(t-1)}) + b_i\right) \quad (3)$$

$$g^{(t)} = tanh\left(D(W_{gx}x^{(t)}) + D(W_{gh}h^{(t-1)}) + b_g\right) \quad (4)$$

$$o^{(t)} = \sigma\left(D(W_{ox}x^{(t)}) + D(W_{oh}h^{(t-1)}) + b_o\right) \quad (5)$$

$$s^{(t)} = f^{(t)} \odot s^{(t-1)} + i^{(t)} \odot g^{(t)} \quad (6)$$

$$h^{(t)} = tanh(s^{(t)}) \odot o^{(t)} \quad (7)$$

$$y^{(t)} = W_{hy}h^{(t)} + b_y \quad (8)$$

where $x_0^{(t)}$ is the vector of raw inputs for the time step $t$, $x^{(t)}$ is the input vector to the LSTM cell, $ReLU$ is the rectified linear unit, $\sigma$ is the sigmoid activation function, $D$ is the dropout operator, $\odot$ denotes pointwise multiplication, $W$'s are network weights, $b$'s are bias parameters, $g^{(t)}$ is the output of the input node, $f^{(t)}$, $i^{(t)}$, and $o^{(t)}$ are respectively the forget, input, and output gates, $s^{(t)}$ represents the states of memory cells, $h^{(t)}$ represents hidden states, and $y^{(t)}$ is the predicted output which is compared to streamflow observations.

The static catchment attributes were concatenated with the meteorological inputs at each time step to produce the input vector. To reduce overfitting, we employed dropout regularization, which stochastically sets some network connections to zero. Here, *D* applies dropout with constant dropout masks to recurrent connections, i.e., the connections that are set to zero stay the same throughout each training instance. This kind of dropout over recurrent connections allows the network to be treated as a Bayesian network (Gal and Ghahramani, 2016). In addition, a nonlinear transformation with a linear function and rectified linear unit (ReLU) was added on the first input layer, following Fang et al. (2020). This was used because without the input transformation layer, some weights of inputs would be directly set to 0 after dropout and lead to information loss. The network outputs one scalar prediction value for each time step, and compares it to the observation for that time step by computing a loss function, which in this case was the root-mean-square error (RMSE) between the observed and predicted discharges. As in Feng et al. (2020a), the Adadelta algorithm, an adaptive learning rate scheme (Zeiler, 2012), was selected as the optimization method for performing stochastic gradient descent on the model parameters of the neural network.

Normalization of inputs and outputs is a useful procedure to facilitate parameter updates by gradient descent. Normally, the loss function is defined over a mini-batch: the model is trained on many basins over the CONUS, and a random subset of hydrographs from some basins are put together to calculate the loss function. In this setup, however, wetter or larger basins contribute more to the loss function than the drier or smaller ones. To prevent this imbalance, we first normalized the daily streamflow by its area and mean annual precipitation to get a dimensionless streamflow, i.e., the runoff ratio, as the target variable. Next, the distributions of daily streamflow and precipitation were transformed to be as close to a Gaussian distribution as possible, using the equation

$$v^* = log_{10}(\sqrt{v} + 0.1) \tag{9}$$

where $v$ is the original value and $v^*$ is the transformed value. Finally, a standard transformation was applied to all the inputs by subtracting the CONUS-scale mean value and then dividing by the CONUS-scale standard deviation. The statistics used for normalization of the test period data were the same as those calculated for the training period data.

There were four hyperparameters: (i) the mini-batch size, which is the number of hydrographs that are put together to calculate the loss function before performing a weight update; (ii) the length of the hydrographs used for training; (iii) the number of hidden units, which is a direct representation of the learning capacity of the LSTM network; and (iv) the dropout probability, which is the probability that a weight is set to 0. As in Feng et al. (2020a), a mini-batch size of 100, an LSTM sequence length of 365, a hidden size of 256, and a dropout rate of 0.5 were selected to run the model. The network training is stochastic in nature. Also similar to the previous setup, all networks in this paper were trained with n = 6 different random seeds. Streamflow predictions resulting from the different random seeds were combined into an ensemble-average prediction. All evaluation metrics were reported for the ensemble-average streamflow, except for the final model transferability experiment (For these experiments detailed in section 2.4.4, we could clearly reach the conclusion from one-random-seed experiments, so there was no need for multiple random seeds). All experiments were implemented using adaptations from the PyTorch library (Paszke et al., 2017), and were performed on an NVIDIA GeForce GTX 1080 Ti GPU.

## 2.2. Basin Datasets

Until now, there had not been a large-scale streamflow benchmark dataset containing extensive basins with reservoirs; CAMELS only has a small fraction of basins with reservoirs. To compile such a dataset, we collected attributes, forcings, and streamflow data for 3557 basins from GAGES-II, which also encompasses most of the CAMELS dataset (see section 2.4). We selected 30 static physical attributes which fit into six categories: (1) basic identification and topographic characteristics, (2) percentages of land cover in the watershed, (3) soil characteristics, (4) geological characteristics, (5) local and cumulative dam variables, and (6) other disturbance variables (see Table S2 in Appendix for more details). Figure 1 plots the location of all 3557 sites and shows five attributes of all basins including slope, forest fraction, soil permeability, normal storage of dams, and freshwater withdrawal. Basin mean forcing data for the period 01/01/1990–12/31/2009 was generated using the same method as for the CAMELS dataset, which was done by mapping a daily, gridded meteorological dataset, Daymet Version 3 (Thornton et al., 2016) to the chosen basin polygons. The Daymet dataset was acquired from the Google Earth Engine (GEE) data catalog (Gorelick et al., 2017) in the form of gridded estimates of daily weather variables for the United States from 01/01/1980 to the present. The basin mean daily time series forcing data were also obtained in GEE using the Map-Reduce functions. Pixels of the gridded data were determined to be in a region according to weighted reducers. Pixels were included if at least 0.5% of the pixel was in the region; their weight was the fraction of the pixel covered by the region. Daily average streamflow was the target variable, for which data for all gauges was downloaded from the USGS website (USGS, 2019). It should be noted that the Daymet data use UTC time (Spangler et al., 2019), while USGS daily values are based on local time (Sauer, 2002). It is difficult to correct this error as they were given in a daily format in the raw data. In this paper, we directly use daily data from the Daymet dataset and the USGS to keep consistent with the CAMELS dataset, as many other studies did. Ideally, one would download sub-daily values from the USGS Instantaneous Values API and shift them to UTC before aggregating to days (or, vice versa, use an hourly forcing product and shift it to local time), as was done in some recent work (Gauch et al., 2020). While we do not think this error changes our conclusions, it calls attention to the need for revisions in datasets like CAMELS.

We also trained and tested models on the CAMELS dataset to allow for comparison to previous results. The CAMELS dataset (Addor et al., 2017; Newman et al., 2015) only included basins which experienced minimal human disturbance, noted as "reference" gages, and excluded basins where human activities including artificial diversions, reservoirs, and other activities in the basin or the channels significantly affected the natural flow of the watercourse (Falcone, 2011).

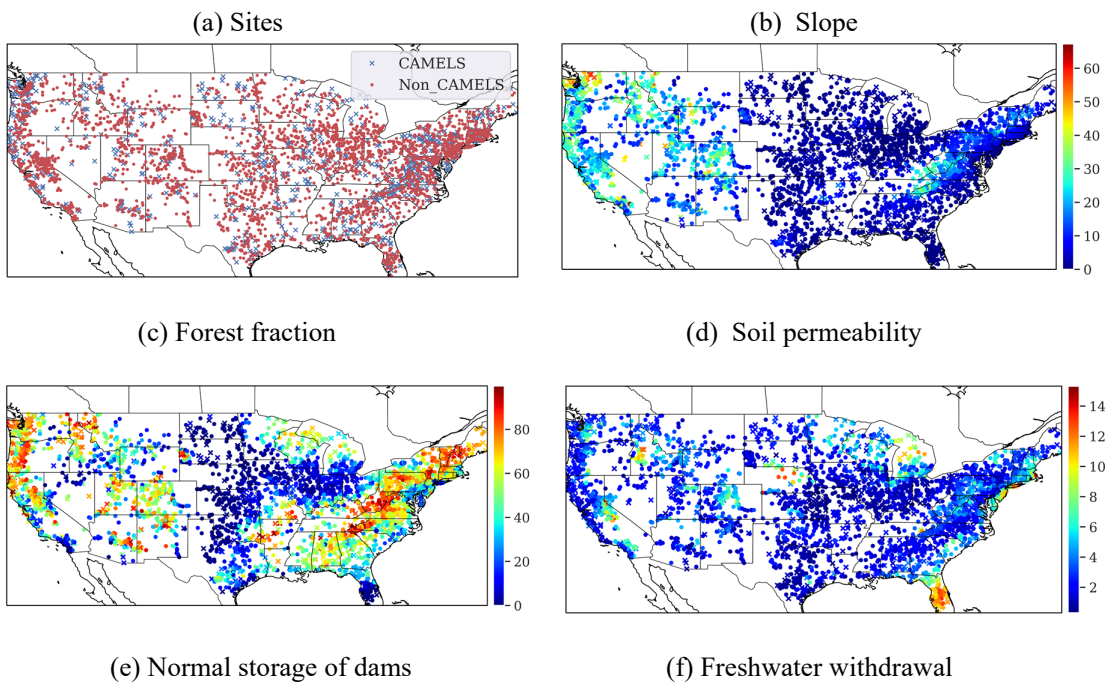

(a) Sites  (b) Slope
(c) Forest fraction  (d) Soil permeability
(e) Normal storage of dams  (f) Freshwater withdrawal

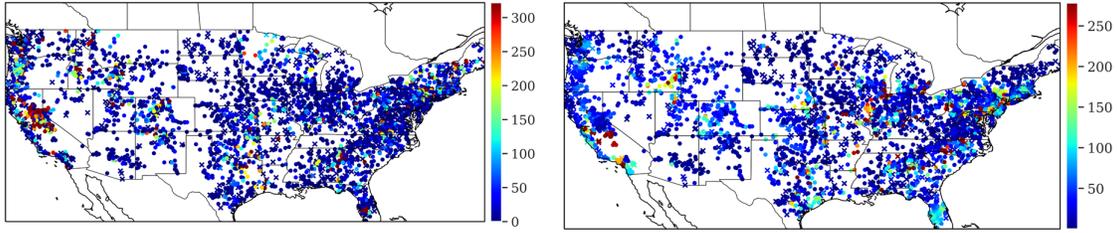

*Figure 1. The location of all 3557 sites and characteristics of the corresponding basins. (a) Locations of all 3557 sites. Blue "x" markers are used to represent sites belonging to the CAMELS dataset, while red "o" points are the other, non-reference sites; (b) Slope: basin mean slope, as a percentage; (c) Forest fraction: percentage of basin with land cover "forest"; (d) Soil permeability: basin average permeability, inches/hour; (e) Normal storage of dams: total normal reservoir storage volume in a basin, megaliters of total storage per sq km; (f) Freshwater withdrawal, megaliters per year per sq km. We excluded some extremely large values of (e) and (f) by choosing values below the 95% percentile value, in order to more clearly show basin diversity.*

## 2.3. Reservoir-related basin characteristics

Degree of regulation (*dor*) refers to the cumulative upstream reservoir storage as a percentage of the average streamflow, and is an important indicator of the impact of reservoirs on streamflow (Lehner et al., 2011). In the present study, it was calculated as the capacity-to-runoff ratio of a basin, defined as follows:

$$dor = \frac{nor}{\bar{q}} \quad (10)$$

where *nor* represents the sum of normal capacity of all reservoirs in a basin (m³ per km²), and $\bar{q}$ is the estimated watershed mean annual runoff, or total volume of water annually leaving the basin via streamflow (m³ per km²), from GAGES-II. A *dor* value of 0.1 was set as the cut-off limit between basins with relatively little human regulation (small-*dor* basins) and basins with relatively large human regulation (large-*dor* basins) based on our preliminary analysis of the distribution of whole-CONUS model's performance across different basins as a function of *dor*. The *dor* is analogous to the commonly used metric of storage ratio (McMahon et al., 2007). A basin with *dor*=0.1 has the approximate storage of about a month of streamflow, which typically would be expected to have significant impact on daily streamflow yet is not enough to heavily modulate flow across seasons. On a side note, *dor* was not the threshold used by CAMELS to select basins. CAMELS contains 344 small-*dor* basins and 32 large-*dor* basins, which represent a much smaller fraction of the CAMELS basins as compared to the overall CONUS.

We hypothesized that reservoir characteristics such as their purposes could be useful. To obtain these attributes, dams listed in the National Inventory of Dams (NID) database (US Army Corps of Engineers, 2018) were spatially joined with the boundary polygons of the basins. To minimize the influences of these differences on our results, we excluded any basins which did not have matching dams included in NID and GAGES-II. Next, for every basin, the sum of the reservoir's normal capacity associated with each dam purpose was calculated. The purpose with the largest associated capacity was considered to be the major purpose of the collective dams in the basin. If there were more than one purpose sharing the largest capacity, we calculated normal storages of these purposes in order of importance (indicated by the order of the letters symbolizing the dam's purposes, e.g. "SC" indicates a primary purpose of water supply followed by flood control), and then chose the most important purpose with the largest capacity. If still more than one purpose was obtained, we treated them as being of equal importance, meaning that there were multiple main dam purposes listed for that basin. There were only a few basins with two categories of main dam purposes (only 1 basin had the main dam purpose of "Debris Control", and only 7 basins had the main dam purpose of "Navigation"), which was not enough to determine statistical characteristics, so they were excluded from the statistical analysis. After all of these processing steps were complete, 656 basins from the 3557-basin dataset were excluded from the statistical analysis in section 2.4.2: 610 basins do not have dams, 38 basins do not have dams listed in either the GAGES-II dataset or NID database, and 8 basins have main dam purposes of "Debris Control" or "Navigation". As a result, 2901 basins with 10 main dam purposes (Table 1) were available to analyze the influence of reservoir types (Table 2).

We added flags to describe the presence of water diversion, based on remarks and comments included in the GAGES-II dataset. "WR_REPORT_REMARKS" reported remarks pertinent to hydrologic modifications from the Annual Data Report (ADR) citation of the USGS, and "SCREENING_COMMENTS" reported screening comments from National Water-Quality Assessment (NAWQA) personnel regarding evidence of human alteration of flow, based on visual (primarily Google Earth) screening. We manually read through the text in these columns, and if there was some description with "diversion" or "divert" for a basin, the presence of diversion for this basin was regarded as "True"; otherwise it was assumed "False". Unfortunately, there was no available data regarding the volume of diversion, and hence diversion could only be used as a qualitative flag for our statistical analysis.

Table 1. Major reservoir purposes for basins in our dam characteristics dataset

| Type | Purpose | Number of Basins |
|------|---------|------------------|
| C | Flood Control and Stormwater Management | 313 |
| F | Fish and Wildlife Pond | 94 |
| H | Hydroelectric | 196 |
| I | Irrigation | 328 |
| O | Other | 163 |
| P | Fire Protection, Stock, or Small Farm Pond | 66 |
| R | Recreation | 1207 |
| S | Water Supply | 426 |
| T | Tailings | 52 |
| X | Unknown | 66 |

## 2.4. Experiments

### 2.4.1. Temporal generalization tests

As we first wanted to determine the level of performance that could be achieved using one model over all 3557 basins in the full dataset (Table 2), an LSTM-based model (LSTM-CONUS) was trained and tested over all of these basins. For comparison to previous studies using the CAMELS dataset, we selected 523 basins (Table 2) from CAMELS (LSTM-CAMELS) to form a training set. The choice of 523 was made for multiple reasons. Firstly, the 3557-basin dataset does not actually contain all of the CAMELS basins. In addition, the attribute data from the GAGES-II dataset and the forcing data used in this study, Daymet Version 3 in GEE (last access in this study: 18 January 2020), were not exactly the same as those used for CAMELS. Finally, by removing some basins with large basin areas, there is a 531-basin subset of CAMELS which has often been selected as the benchmark set for rainfall-runoff modeling in previous work (Feng et al., 2020a; Kratzert et al., 2019b). An intersection between the 3557 basins and this 531 benchmark CAMELS subset basins resulted in the 523-basin "baseline" CAMELS dataset we used here. All models were trained using data from 1 January 1990 through 31 December 1999, and testing was done using data from 1 January 2000 through 31 December 2009.

### 2.4.2. Exploring the impacts of reservoir attributes on model performance

There are many reservoir attributes that could potentially inform improvements in streamflow modeling, such as dam storage or distance from gage location to dam. As the first paper (to the best of our knowledge) to study continental-scale streamflow prediction in dammed basins in a deep learning context, we explored the impacts of multiple reservoir attributes and anthropogenic factors (details in Appendix Figure S2). Then, within the scope of this paper and partially consistent with McManamay (2014), we examined three major factors having significant influence on our model performance: capacity-to-runoff ratio (degree of regulation, *dor*), main dam purpose, and presence of diversion. As the models utilized in this study were basin-centric, these factors needed to be aggregated to each basin, which was done following the procedures discussed in Section 2.3.

Table 2. Datasets used in the this study

| Name | Number of basins | Explanation |
| --- | --- | --- |
| full dataset | 3557 | Basins with complete streamflow records during 1990/01/01-2009/12/31, selected from GAGES-II (section 2.4.1) |
| 523-CAMELS dataset | 523 | Basins contained both in full dataset and CAMELS (section 2.4.1) |
| dam characteristics dataset | 2901 | Subset of full dataset, containing basins used to explore the impacts of the three factors: capacity-to-runoff ratio (*dor*), dam purpose, and diversion (section 2.4.2) |
| zero-*dor* dataset | 610 | Subset of full dataset, containing basins without dams (section 2.4.3, 2.4.4) |
| small-*dor* dataset | 1762 | Subset of full dataset, containing basins with 0 < *dor* < 0.1 (section 2.4.3, 2.4.4) |
| large-*dor* dataset | 1185 | Subset of full dataset, containing basins with *dor* ≥ 0.1 (section 2.4.3, 2.4.4) |

### 2.4.3. Stratification by reservoir regime vs. pooling data together

For DL models in general, providing more data often leads to model improvements. From the perspective of machine learning, then, lumping all data together would thus seem to be the obvious procedure to follow, given the likely beneficial impacts on modeling performance as well as simple implementation. However, it remains possible that stratification by reservoir attributes might result in clear separation basins with different latent (unknown) attributes. Hence, our research question 2 raised in the Introduction became two sub-questions: (2A) Should we group all basins together, or classify basins into certain types and train models for each class separately to achieve the best performance? (2B) Do basins with varied reservoir regimes (no reservoir, small reservoir, or large reservoirs) function fundamentally differently? This could be proven true if basins trained in one regime cannot apply to basins in another regime.

To answer question 2A, all basins in the full dataset were divided into three groups (Table 2): zero-*dor* basins (*dor*=0), small-*dor* basins (0<*dor*<0.1) and large-*dor* basins (*dor*≥0.1). We trained models on these different groups individually, as well as together in various combinations. First, we trained and tested three LSTM-based models, called LSTM-Z, LSTM-S, and LSTM-L (we used "LSTM-x" to represent the LSTM-based models, which was different from the naming method for the datasets), on zero-*dor*, small-*dor* and large-*dor* basins, respectively. Second, basins from two of the three groups were combined into training sets for three additional LSTM-based models: LSTM-ZS (trained on zero-*dor* and small-*dor* datasets), LSTM-ZL (trained on zero-*dor* and large-*dor* datasets), and LSTM-SL (trained on small-*dor* and large-*dor* datasets), but these three models were tested on basins from each of zero-*dor*, small-*dor*, and large-*dor* datasets. Finally, the testing results of basins in these three groups were compared to results for the same basins from the LSTM-CONUS (trained on full dataset) model.

### 2.4.4. Model transferability experiments

To answer question (2B) raised in 2.4.3, we ran a set of predictions in ungauged basins (PUB) experiments, in which models trained in one set were tested in other sets. Further, when a model is trained in some basins and tested in others, the performance will naturally degrade. Therefore, we added control experiments where models were trained and tested on the same categories of basins, which helped to disentangle the effects of reservoir regime and spatial extrapolation.

For example, zero-*dor* basins were divided into two batches (Train-z and PUB-z) with a ratio of 1:1 for training and test, respectively. We ensured that each of these cases was representative of the full group by including basins from every LEVEL-II ecoregion (Omernik and Griffith, 2014). The model trained on the Train-z set is then tested on Train-z itself, PUB-z and a subset (PUB-s) of the small-*dor* basins. These three test sets represent temporal generalization alone, spatial extrapolation and "spatial extrapolation+difference in reservoir regime", respectively. Similarly, we separated the small-*dor* dataset into Train-s and PUB-s, and the large-*dor* dataset into Train-l and PUB-l. We also ran experiments with a mixed training set, e.g., Train-z and Train-s were merged to form one training dataset called Train-zs. Once trained on Train-zs, the LSTM-based model was tested individually on PUB-z and PUB-s. Two more training sets, combining zero-*dor* basins with large-*dor* ones (Train-zl), and pairing small-*dor* basins with large-*dor* ones were set up in the same way (Train-sl). It was not practical to attempt all possible combinations, but the combinations used sufficiently answered the question (2B).

Finally, a fourth sub-experiment was added for comparison, to test the transferability of the LSTM-based model trained on the 523-CAMELS dataset. The basins of the 523-CAMELS dataset were also divided into the training (Train-c) and test (PUB-c). Then, the models trained on Train-c were tested on itself and other subsets (PUB-c/PUB-z /PUB-s/PUB-l). The details of all four of these sub-experiments are listed in Table 3.

Table 3. A summary of the training and testing datasets for sub-experiments exploring PUB with dams. All models were trained from January 1990 through December 1999, and tested from January 2000 through December 2009. Multiple basin counts are given for each case of the first three sub-experiments, as we ran two tests (and therefore performed the basin groupings twice) for each case. For example, in the first sub-experiment, Train-z had 299 basins for the first run, and 309 basins for the second run. We list the Train-z and PUB-z datasets twice in the first and second sub-experiments, because they belong to two independent sub-experiments.

| sub-experiment ID | training dataset (explanations) | test dataset (explanations) |
|---|---|---|
| 1 | Train-z (299/309 randomly selected zero-*dor* basins) | Train-z (same as the training set) |
| | | PUB-z (309/209 zero-*dor* basins that are different from those in Train-z) |
| | | PUB-s (300/292 randomly selected small-*dor* basins) |
| | Train-zs (A mixture of 544/560 zero-*dor* or small-*dor* basins) | PUB-z (280/272 zero-*dor* basins that are different from those in Train-zs) |
| | | PUB-s (280/272 small-*dor* basins that are different from those in Train-zs) |
| 2 | Train-z (295/305 randomly selected zero-*dor* basins) | Train-z (same as the training set) |
| | | PUB-z (305/295 zero-*dor* basins that are different from those in Train-z) |
| | | PUB-l (297/289 randomly selected large-*dor* basins) |
| | Train-zl (A mixture of 512/528 zero-*dor* or large-*dor* basins) | PUB-z (264/256 zero-*dor* basins that are different from those in Train-zl) |
| | | PUB-l (264/256 large-*dor* basins that are different from those in Train-zl) |
| 3 | Train-s (871/879 randomly selected small-*dor* basins) | Train-s (same as the training set) |
| | | PUB-s (879/871 small-*dor* basins that are different from those in Train-s) |
| | | PUB-l (639/634 randomly selected large-*dor* basins) |
| | Train-sl (A mixture of 876/888 small-*dor* or large-*dor* basins) | PUB-s (444/438 small-*dor* basins that are different from those in Train-sl) |
| | | PUB-l (444/438 large-*dor* basins that are different from those in Train-sl) |
| 4 | Train-c (257/264 basins in the 523-CAMELS dataset) | Train-c (same as the training set) |
| | | PUB-c (264/257 basins that are different from the Train-c dataset, but still in the 523-CAMELS dataset) |
| | | PUB-z (383 zero-*dor* basins that are different from the 523-CAMELS dataset) |
| | | PUB-s (1482 small-*dor* basins that are different from the 523-CAMELS dataset) |
| | | PUB-l (1169 large-*dor* basins that are different from the 523-CAMELS dataset) |

## 2.5. Metrics

In this study, the metrics used to mathematically quantify the accuracy of the models included bias, Pearson's correlation (Corr), the Nash-Sutcliffe model efficiency coefficient (NSE) (Nash and Sutcliffe, 1970) and Kling-Gupta efficiency (KGE) (Gupta et al., 2009). Bias is the mean difference between modeled and observed values. Corr is the linear correlation coefficient between modeled and observed values, and is not influenced by bias. NSE is a normalized statistic that determines the relative magnitude of the residual variance compared to the measured data variance. KGE is a nonlinear combination of correlation, flow variability measure, and bias; it is another common metric to evaluate how well the models perform. We also reported the percent bias of the top 2% high flow volume range (FHV) and the percent bias of the bottom 30% low flow volume range (FLV) (Yilmaz et al., 2008). FHV and FLV highlight the performance of the model for peak flows and baseflow, respectively. Metrics for all experiments in this study are reported for the test period (01/01/2000-12/31/2009).

## 3. Results and Discussion

### 3.1. CONUS-scale model with reservoirs

For the 3557 basins in the full dataset, the ensemble median NSE of the CONUS-scale model reached 0.74 (Figure 2c, details of ensemble experiments recorded in Appendix Table S3). This value is at the same level as the previous benchmarks with the CAMELS reference-basin dataset (Feng et al., 2020a; Kratzert et al., 2020), despite that 83% of the 3557 basins have dams present in GAGES-II. When the models trained on CAMELS (LSTM-CAMELS) and CONUS (LSTM-CONUS) were tested on the 523-CAMELS baseline reference dataset, both achieved a median NSE values of 0.75 (Figure 2c, more details in Appendix Table S3).

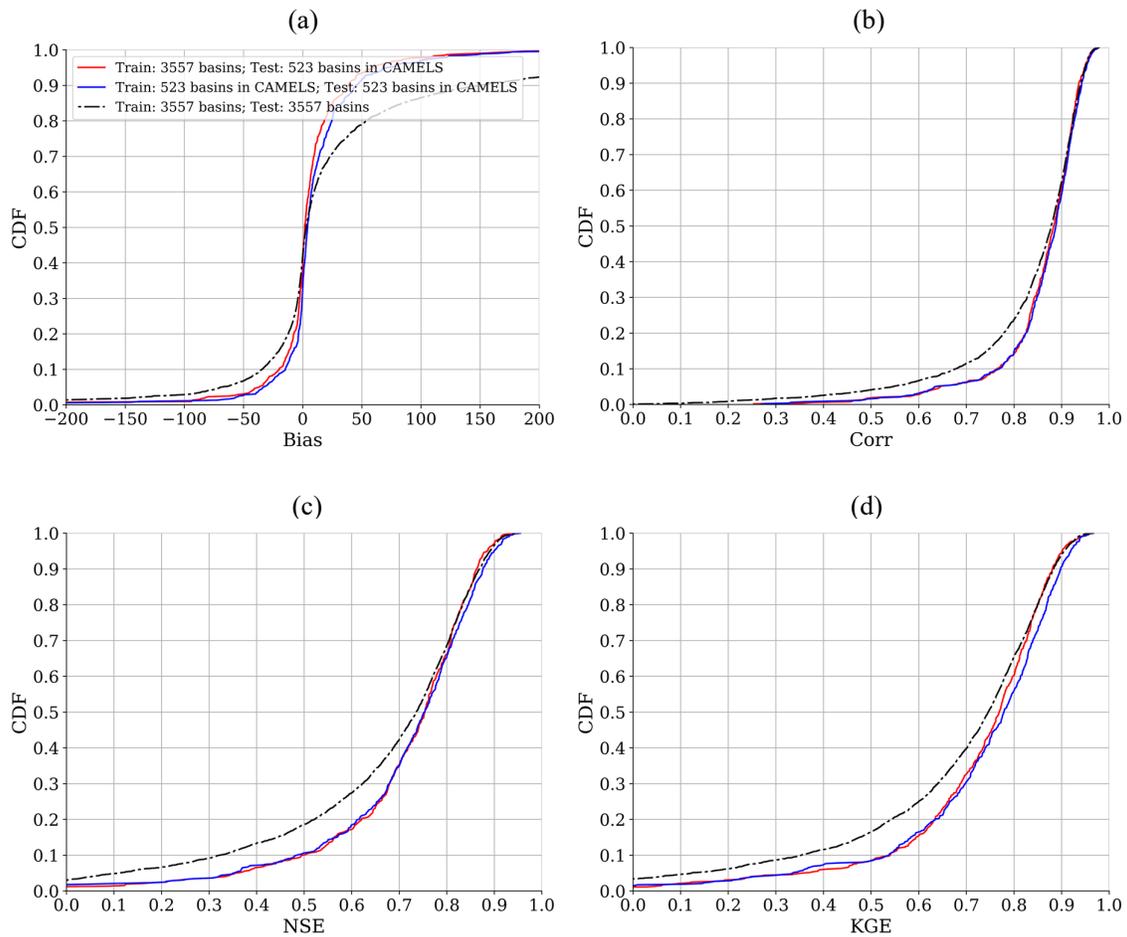

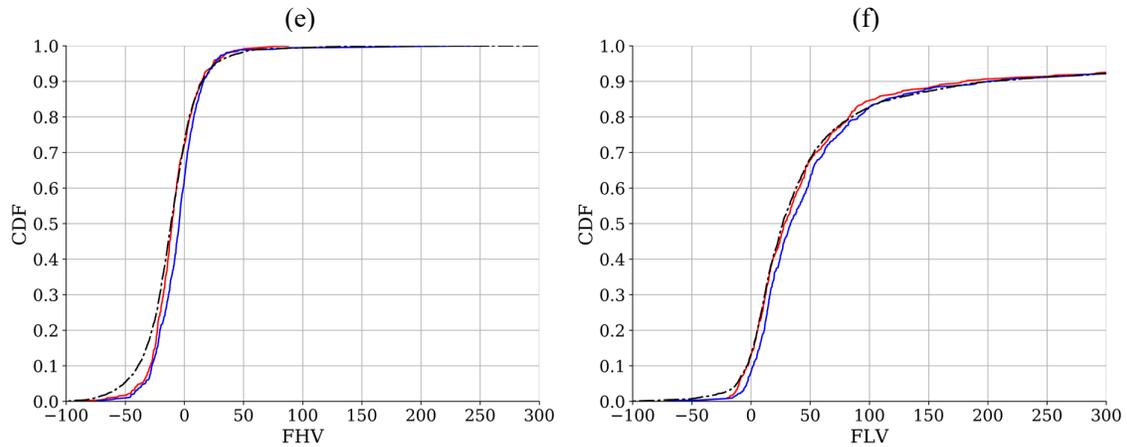

*Figure 2. Comparison of the empirical cumulative distribution functions (CDF) for the 523 basins tested in LSTM-CONUS and LSTM-CAMELS, and the 3557 basins in LSTM-CONUS. The CDF of FLV does not reach 1.0 because the 30% low flow interval for some basins is completely composed of zero-flow observations. Therefore, for these basins, the percent bias is infinite, and thus the x-axis cannot include them.*

The high NSE for the entire set was somewhat unexpected, because we had earlier thought that reservoirs would create challenges for LSTM and there may not be reliable mapping relationships that could be learned on a large scale. Comparing our results to those reported in the literature, a NSE of 0.74 certainly represents a state-of-the-art prediction for basins with reservoirs, and a much more operationally-reliable model. Besides the values reported in literature summarized in the Introduction and Table S1, many of which reported negative NSEs for this challenging problem, the closest value we can find in the literature was Payan et al. (2008), who added reservoirs into a simple lumped hydrologic model, tested this model in 46 basins (mostly in France), and reported a mean NSE of 0.68. We would also like to note that the meteorological data for CONUS seems to have larger error than the European counterpart, which could lead to our model presenting an even higher NSE with European basins if we were to train our models there. In line with this hypothesis, some of our previous work showed that we could obtain a NSE of 0.84 for CAMELS-GB (Coxon et al., 2020), which has 670 basins from United Kingdom (Ma et al., 2021), while the same model with the same training procedure could only achieve a NSE of 0.74 for CAMELS over CONUS.

When tested on the 523-CAMELS dataset, the expanded dataset led to slightly improved overall bias with almost the same correlation but slightly decreased KGE (noticeable by comparing red and blue lines in Figure 2a-b,d). Since KGE is a composite metric of correlation, flow variability, and bias, we suspect that additional samples in the larger dataset enlarged the flow variability, which makes it a little more difficult for LSTM-CONUS to capture the flow variability for the 523 basins. This hypothesis can be further validated by looking at the values for FHV and FLV. The median FHV values when tested on the 523 CAMELS basins were -10% for LSTM-CONUS and -4% for LSTM-CAMELS, showing a minor increase in high-flow bias for the expanded dataset (Figure 2e). In contrast, for the same test set, the low-flow simulations were improved by the use of a bigger training dataset, as the median FLV values were 28% for LSTM-CONUS, and 33% for LSTM-CAMELS (Figure 2f). Compared to CAMELS, we suspect the expanded set may contain a higher fraction of basins with large reservoirs which attenuate the peak flow, and hence the LSTM-CONUS model tended to predict lower peaks.

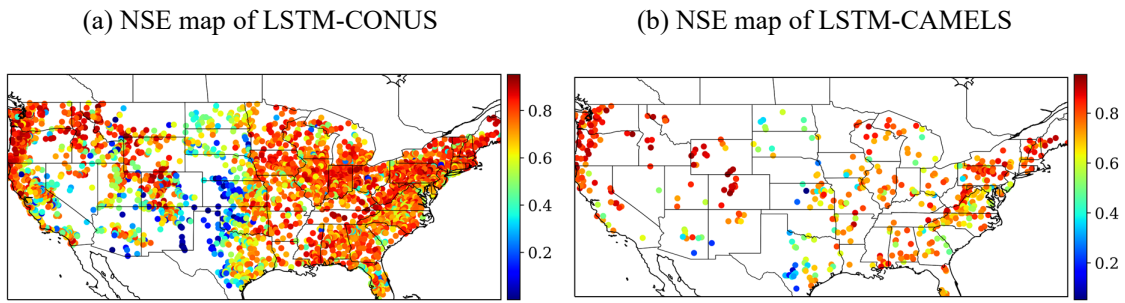

Figure 3. NSE spatial patterns of the ensemble results of (a) LSTM-CONUS and (b) LSTM-CAMELS.

LSTM-CONUS and LSTM-CAMELS both showed good performance in the northwestern CONUS and most parts of the eastern CONUS, but had relatively poor performance on the Great Plains, Texas, Oklahoma, Kansas, and parts of California (Figure 3). The regional distribution of NSEs is largely in line with earlier work (Feng et al., 2020a), where basins on the Great Plains and the extremely-dry southwestern border performed poorly with LSTM-based modeling. Evidently these basins in the central CONUS continue to pose challenges for LSTM despite the larger dataset, perhaps because they are still large basins where the homogeneous assumption of the LSTM-based models breaks down.

**3.2. Analysis of the impacts of reservoir-related factors**

Using the results from the CONUS-scale simulation (LSTM-CONUS), we explored the uncertainty of the current LSTM-based model guided by three attributes: the capacity-to-runoff ratio (degree of regulation, *dor*), the purpose of the dam and its associated reservoir, and the presence of diversion (Figure 4a). There was a clear pattern regarding *dor*: regardless of the purpose, the overall model performance, as quantified by the median NSE, was always better for small-*dor* basins than for larger-*dor* ones (see Figure 4d). This observation differs from previously-reported results obtained with a process-based model (Shin et al., 2019), which had more difficulty predicting the streamflow of basins with small-capacity reservoirs (corresponding to small *dor*). The management policies of reservoirs could change over time and we think that is potentially the reason why the model did not perform as well for large-*dor* basins. However, for small-*dor* reservoirs, the model still delivered excellent performance so such changes in policies may not have resulted in dramatic impacts for these small reservoirs. A first-order visualization of the impacts of other control variables are given in Appendix Figure S2.

Exploring model uncertainty based on dam purpose not only showcased the uncertainty of the LSTM-based models, but also clearly indicated that different types of reservoirs exert varied influences on streamflow. Among all the various dam purposes, basins with reservoirs mainly for recreation (R) or water supply (S) were easier to model. It may be inferred that the water storages of these reservoirs changed relatively little on a daily scale to achieve their purposes and therefore had less impact on the streamflow than other reservoirs (Ryan et al., 2020). Three types of reservoir purposes stood out as being more challenging to predict (Figure 4b): fire protection or farm ponds (P), irrigation (I), and hydroelectric (H). Basins with "P" reservoirs, for any *dor* value range regardless of the presence of diversion, were difficult to predict and had the worst performance of all those in the small-*dor* category. This indicates that LSTM had trouble finding a universal relationship to model processes for a chain of many small, individually-regulated ponds. Difficulty in modeling irrigation reservoirs was not unexpected, as it has been shown that irrigation water usage has specific seasonal variations, and is related to the crop type, field, and other site-specific information (Shin et al., 2019). Critical information that would help with modeling for these basins, such as water use and timing, is not generically available. Likewise, the operational policies of hydroelectric (H) dams seek to optimize electricity production, and are therefore influenced by the prices on the local electricity grid (Giuliani et al., 2014), which were not included in this dataset.

The presence of diversion substantially decreased NSE values (Figure 4a). For instance, it is visibly apparent that there are smaller NSE values for dam purposes "I", "O", "P", and "R" in the basins with diversion. This was also expected: diversion influences the water balance, but because no information about the quantity of diverted water was available to the LSTM-based model, the model couldn't understand the imbalance, leading to reduced prediction performance. A clearer separation is seen in the results of four specific cases, which differ by combinations of only two categorical variables -- the *dor* value range, and the presence of diversion (Figure 4c). The median NSEs for small-*dor* basins without diversion, small-*dor* basins with diversion, large-*dor* basins without diversion and large-*dor* basins with diversion were 0.78, 0.76, 0.65, and 0.62, respectively. It was evident that LSTM could reach the best performance in small-*dor* basins without diversion, while the worst performance occurred in large-*dor* basins with diversion, and thus the effects of the two factors seem to be additive.

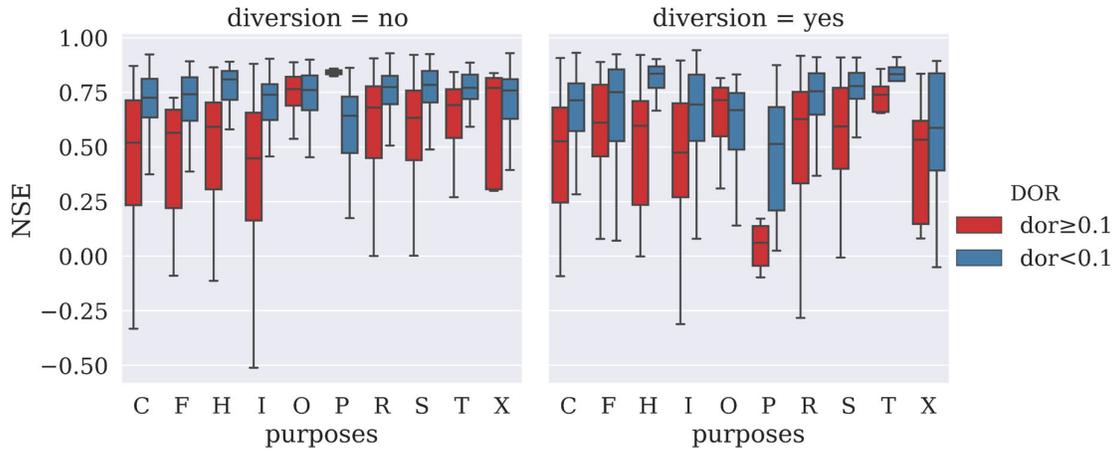

(a)

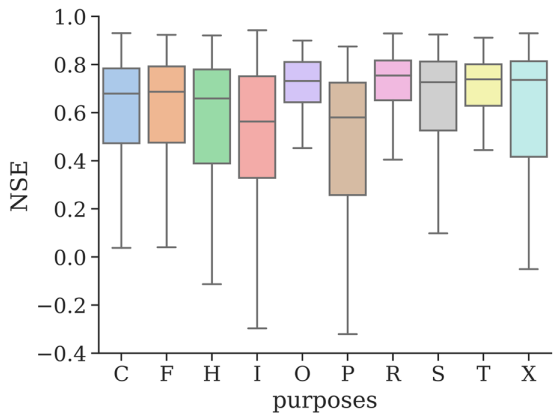

(b)

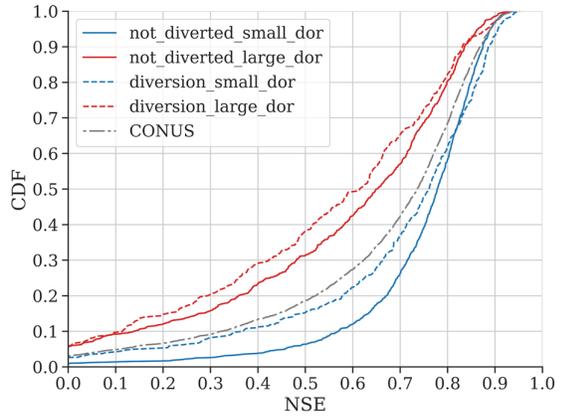

(c)

(d)

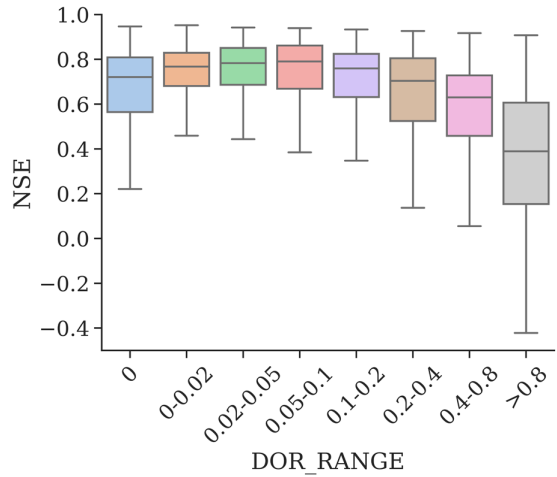

*Figure 4. (a) NSE distributions with three categorical variables: dor value range ("small-dor" basins have 0<dor< 0.1 and "large-dor" basins have dor ≥ 0.1), main purposes of reservoirs in a basin, and presence of diversion. Dam purposes are C: Flood Control and Stormwater Management; F: Fish and Wildlife Pond; H: Hydroelectric; I: Irrigation; O: Other; P: Fire Protection, Stock, or Small Farm Pond; R: Recreation; S: Water Supply; T: Tailings; and X: Unknown. (b) NSE distribution for basins with different main dam purposes. (c) NSE empirical cumulative distribution function curves from LSTM-CONUS and four cases resulting from combinations of two categorical variables: dor range and presence of diversion. The blue and green lines respectively represent the NSE distributions of small-dor basins with and without diversion, which were picked out from the ensemble result of LSTM-CONUS. The red and orange lines respectively indicate the NSE distributions of large-dor basins with and without diversion. The grey dashed line represents the empirical CDF of LSTM-CONUS. (d) NSE as a function of dor values all 3557 basins; the ranges of dor values: 0, (0, 0.02], (0.02, 0.05], (0.05, 0.1], (0.1, 0.2], (0.2, 0.4], (0.4, 0.8], >0.8, where "()" means a left side half open interval; the correspond numbers of basins in each range: 610, 1076, 377, 309, 311, 277, 247, 350; other plots in this figure are for dam characteristics dataset shown in table 2.*

The main challenges for LSTM-based modeling of reservoirs are clearly delineated (Figure 4a): LSTM had difficulty predicting streamflow for large-*dor* basins with dams for fish and wildlife, flood control, hydroelectric power generation, irrigation, and fire protection, with difficulty increasing in this order. Diversion further added to the challenge. To our knowledge, such identification of specific challenges has not been previously reported. Additionally, it was not previously clear that these challenges mainly exist only for large-*dor* basins. Small-*dor* basins, even those with reservoirs for irrigation and hydroelectric purposes, can be reasonably captured by LSTM, presumably because they have limited adaptive capacity. LSTM can approximate an optimal information extractor, which suggests that we did not supply sufficient information needed to model the more challenging cases and provides a targeted direction for future work.

*dor* is apparently a major control on LSTM model performance (Figure 4d). Interestingly, small-*dor* basins, instead of zero-*dor* basins, have the highest performance. The median NSE in the 0.05-0.1 *dor* bin is almost 0.8, a very high number (we offer explanations later). Below *dor*<0.1 human decisions cannot shift water availability across seasons. As discussed earlier, basins with *dor*=0.1 have the reservoir storage equivalent to approximately one month of average streamflow. As *dor* gets bigger than this amount, they have more capability to regulate flow on a seasonable scale, and the impact of human choice becomes more prominent. We also found the basin with more reservoirs could have equivalent or higher performance (Figure S2l), which suggests the difficulty may have mainly come from one or few largest dams. Due to sometimes unpredictable human decisions influenced and also the nonstationarity in such decisions, e.g., shift in reservoir management policies, the *dor*>0.1 becomes increasingly difficult to simulate. This figure is also the basis for us to choose *dor*=0.1 as the threshold. Despite the challenges for large-*dor* basins, we nonetheless note that even for these basins, LSTM obtained a median NSE of 0.65 for basins without diversion, which is higher than many literature values reported in Table S1. To put things even further into context, a recent study for a basin with a major dam (USGS 11462500, Russian River near Hopland, California, *dor* = 0.17) reported oftentimes negative daily NSE values and correlation between 0.5 to 0.8 for different months of the year (Kim et al., 2020). In contrast, the CONUS-scale model developed in this study reported a very high NSE value of 0.88 and correlation of 0.94 for this specific station. For a different comparison, the National Water Model reported an NSE of 0.62 for reference basins in CAMELS (Kratzert et al., 2019a).

### 3.3. Impacts of training dataset

Our experimental results suggest that datasets with different *dor* value ranges can be trained together to enhance overall performance, and at the very least, grouped training should not exert a significant detrimental impact on the model (Figure 5a, see more details in Tables S3 and S4, Appendix). With the inclusion of small-*dor* basins in the training set (LSTM-ZS), there was a small improvement in predictions for undammed basins (Wilcoxon signed-rank test: $p=4.9 \times 10^{-6}$). For small-*dor* basins, there were no clear differences in test performance when training with zero-*dor* basins together. In the large-*dor* basins, as compared to the result of LSTM-L (training with only large-*dor* basins), all other cases reported slightly increased NSE values and fewer "catastrophic failures" (cases with NSE close to or smaller than 0), suggesting that new information was brought in by pooling information together. It is possible that the inclusion of zero-*dor* or small-*dor* basins allowed the model to better understand natural flows and enabled better modeling of the large-*dor* basins. Such a pattern fits with our general observations obtained from training DL models.

We did see a slight exception to this pattern, however, when adding large-*dor* basins to the training set. When large-*dor* basins were added to the training set, a minute deterioration in NSE was observed when this model was tested on zero-*dor* and small-*dor* basins: the median NSE decreased from 0.72 to 0.71 for LSTM-ZL (left panel of Figure 5a, Wilcoxon signed-rank test: $p=1.3 \times 10^{-4}$), and there was a declination from 0.79 to 0.78 shown for LSTM-SL (center panel of Figure 5a, Wilcoxon signed-rank test: $p=1.2 \times 10^{-32}$). We hypothesize that operations of large reservoirs are characteristically different from those of smaller reservoirs, and therefore the inclusion of large reservoirs introduced some noise to the data and made it more difficult for LSTM to grasp a universal pattern. Nevertheless, the adverse impact was quite minor. This result, along with our other observations of LSTM-CONUS (Section 3.1), also imply that it should be possible to fine-tune the LSTM-CONUS model for a local region to obtain refined simulations.

We were surprised to see that small-*dor* basins had notably higher NSE values (median NSE ~0.79) than zero-*dor* basins (median NSE ~0.72) (Figure 5a). Two hypotheses could potentially explain this phenomenon: first, that the small-*dor* basins may be concentrated in certain areas, e.g., mountainous areas, where NSEs tend to be higher; second, that a small-*dor* reservoir may serve as a buffer to boost the storage of the system, thereby reducing the impacts of flash precipitation peaks which are challenging to model (Feng et al., 2020a). Looking at the basins on a map and in the parameter space (Figure 5b), however, while mountainous basins do have higher NSEs, the zero-*dor* and small-*dor* basins are mixed in space and there is no spatial aggregation of one or the other. Therefore, we reject the first hypothesis (concentration) and lean toward the second one (buffer).

(a)

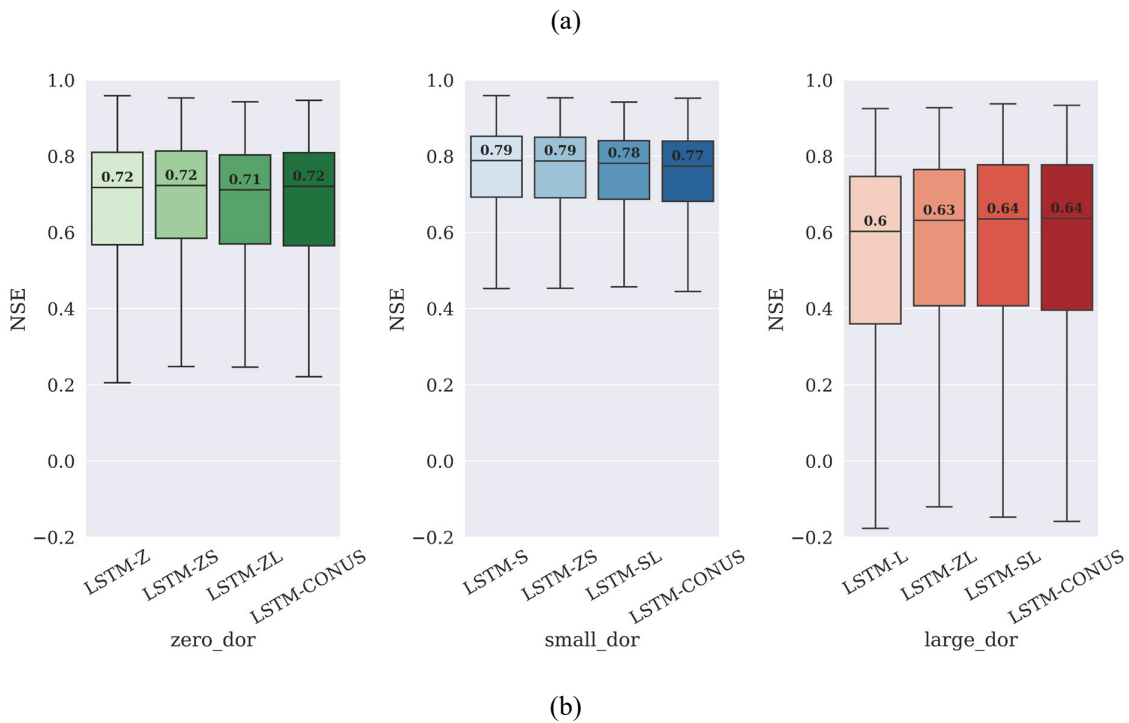

(b)

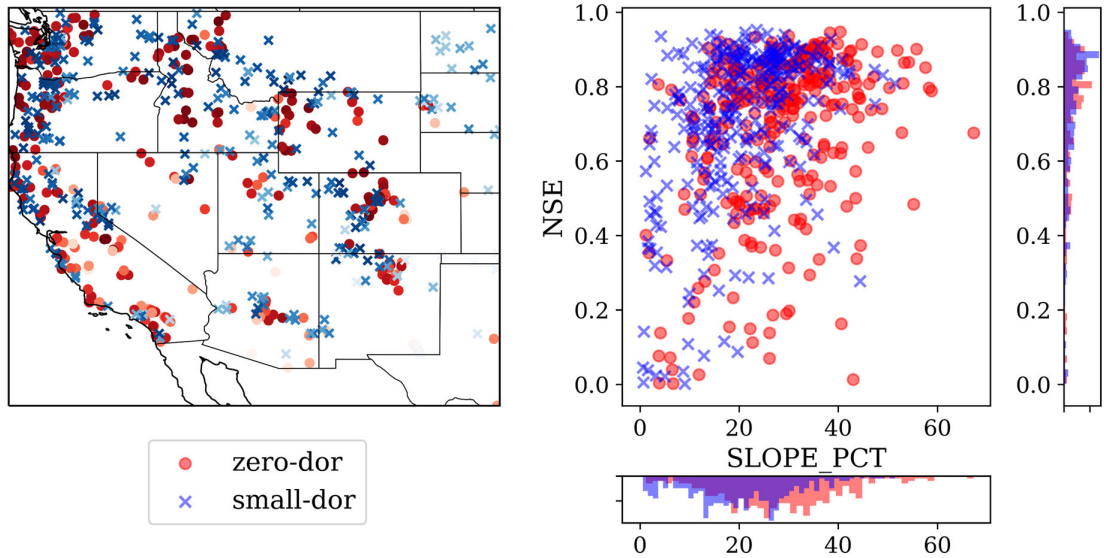

*Figure 5. (a) Boxplots of NSE values for zero-dor basins (Z, dor=0), small-dor basins (S, 0<dor<0.1) and large-dor basins (L, dor≥0.1). Green, blue, and red boxes show the results from models respectively tested on zero-dor, small-dor, and large-dor basins, while the training sets are noted on the x-axis labels. For each color, the lightest-colored box was trained solely with the same subset of basins on which it was tested, while the others had additional subsets included in the training sets. Basins in the test sets were always subsets of the training sets, and the models were trained in 1990-1999 and tested in 2000-2009. (b) The left part is a NSE map of the western CONUS where small-dor and zero-dor basins coexisted. There are 303 zero-dor basins and 310 small-dor basins shown here. The right is a scatter plot of the relationship between NSE and SLOPE_PCT (mean watershed slope, as a percent). The NSE values are part of the results for LSTM-CONUS (section 3.1). Red circular markers represent the zero-dor sites, and blue x-shaped markers represent the small-dor sites. For the map only, sites with lighter colors have lower NSE values.*

Additionally, we were also surprised to see that LSTM showed reasonably good performance on even large-*dor* basins, with median NSE values of ~0.64 in the overall CONUS training sets (the rightmost boxplot in Figure 4a), respectively, which were still comparable to SAC-SMA's median NSE of 0.65 (Feng et al., 2020a) for reference basins. This result suggests a large advantage of LSTM for modeling reservoirs as compared to earlier methods.

### 3.4. The PUB experiments and model transferability

As we asked in question 2 in the introduction, were the NSE values for dammed basins similar to previous results with CAMELS because these basins in fact behave similarly? If this was not the case, how different are these basins? Our stratified PUB experiments showed that there were substantial differences between zero-*dor*, small-*dor*, and large-*dor* basins such that applying models trained only on one type of basin to other basin types caused significant performance drop that could not be explained solely by spatial extrapolation (Figure 6). For example, the median NSE values for "Train-z", "PUB-z", and "PUB-s" were 0.65, 0.51, and -0.06, respectively (Figure 6a). The scenario Train-z was a temporal test only, so this NSE value of 0.65 represents model performance without spatial extrapolation (this value was lower than LSTM-Z shown in Figure 5a because the training sample size was smaller: the zero-*dor* basins were randomly split for this experiment, as explained in section 2.4.4). The decline from 0.65 to 0.51 for PUB-z was then due to spatial extrapolation in the same zero-*dor* group. The more dramatic decline from 0.51 for PUB-z to -0.06 for PUB-s can be entirely attributed to the behavioral difference between zero-*dor* and small-*dor* basins. We also note larger declination for large-*dor* basins (Figure 6b-c), with median NSE values of -0.19 and 0.18 for the PUB-l cases.

Including diverse basins in the training dataset substantially elevated overall PUB performance. The mixed training sets (Train-zs, Train-zl, and Train-sl, the boxes on the right side of each panel in Figure 6a-c) had greatly improved median NSE values, as well as greatly reduced incidences of catastrophic failures (cases with NSE close to 0).

It is noteworthy to mention that when we trained a model solely on basins subset from the 523-CAMELS dataset and then tested it on the other basins of 523-CAMELS as well as zero-, small-, and large-*dor* basins, the model gave outright disastrous results for PUB-z, PUB-s, and PUB-l (Figure 6d). This means that CAMELS basins, as they are reference basins, differ fundamentally from the others, even from the zero-*dor* basins. This result distinctively highlights the danger of using CAMELS basins as the whole training set for continental-scale modeling, and also suggests we cannot simply ignore small reservoirs or simply treat them as being equivalent to reference basins.

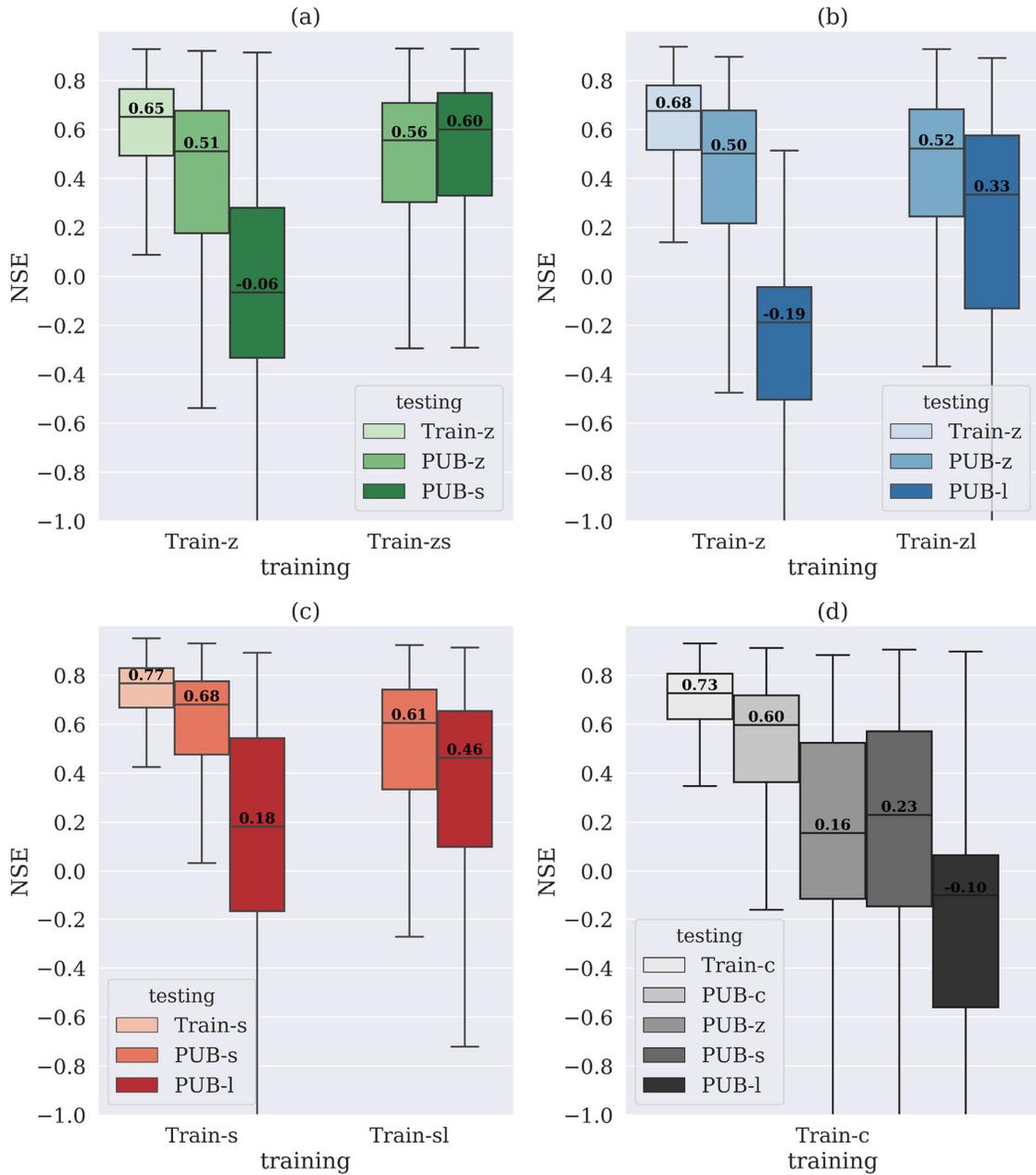

*Figure 6. Boxplots from PUB sub-experiments where training and testing basins were from different combinations of basin types: c indicates 523-CAMELS, z indicates zero-dor basins, s indicates small-dor basins, and l indicates large-dor basins. Combinations of letters indicate that a combination of the indicated basin types were used (refer to Table 3 for details). The drop in performance from training basin-located test results to PUB-basin-located test results of the same type (e.g. Train-z vs PUB-z) represents the effect of spatial extrapolation, while the drop across different basin type combinations (e.g. PUB-z vs PUB-s) represents the effect of migrating models across reservoir regimes. A side note: the PUB-c in (d), with a median of 0.60, is not comparable to other PUB tests in the literature. Here we only used ~260 CAMELS basins as training data and did not employ an ensemble for different random seeds (so as to be inline with other experiments in this figure). This test is solely shown to highlight the difference between the CAMELS basins and the others.*

### 3.5. Further Discussion

In future work, we could allow LSTM to estimate model uncertainty based on input attributes, as shown in the modeling of soil moisture (Fang et al., 2020) and rainfall-runoff (Klotz et al., 2020). To further improve modeling capabilities for the more challenging cases, it could be useful to incorporate more information regarding water use, electricity price patterns, and estimated diversion rates from sources like water management models (Yates et al., 2005) into the context of optimization processes (Giuliani et al., 2016). Fine-tuning may be another approach to improve predictions in more challenging basins (Sampson et al., 2020). For example, Ma et al. (2021) transferred their model trained on the CAMELS basins over to a few basins in Sichuan province in China and obtained better results than the model trained with all local basins. Other reservoir-related information such as distribution of the storage capacity among the basin's reservoirs, surface water area, or storage change in a basin may also be used as inputs through an encoder unit (Feng et al., 2020b). Moreover, physics-guided machine learning (Read et al., 2019) could be employed to provide more stability where monitoring data is scarce. In addition, a distributed version of the deep learning models could represent the spatial heterogeneity of a basin and may perform better than the lumped ones for large basins. In the future, machine-learning-based routing schemes (Bindas et al., 2020) can be added to support flood modeling in major rivers.

As a rule of thumb for DL models, pooling data together almost always helped improve modeling, which was confirmed by the zero-*dor* and small-*dor* cases shown in this study. However, here the large-*dor* basins could slightly pull down the metrics for other cases, which deviated, albeit in a minor way, from this rule. We think that this was due to a combination of the rainfall-runoff processes from different basins having very dissimilar patterns, and the information from the inputs not being enough to discern differences between reservoir regimes, causing the LSTM-based model to struggle in fitting all of this information into one universal model. We suspect that the large-*dor* basins represent an extreme case of the problem of unmodelable dissimilarity in geoscience. The cut-off *dor* of 0.1 in this paper is an operational threshold, but may not be the only choice. Other *dor* cut-off values may also be applicable, but this was not the focus of this paper. Future work should concentrate on how to incorporate more information and tune the model structure to train a universal model for all non-regulated/regulated basins.

## 4. Conclusion

Prior work has documented the success of modeling rainfall-runoff processes with LSTM in reference basins with minimal anthropogenic impacts. However, to our knowledge, no previous deep-learning based study focused on basins significantly impacted by reservoir operations at a continental scale, or the modeling implications of reservoir attributes. For this work, we created a new dataset consisting of 3557 basins over the CONUS, and trained an LSTM-based model which achieved an ensemble test median Nash Sutcliffe model efficiency coefficient (NSE) of 0.74. This performance was at the same record level as reported for previous LSTM-based modeling benchmarks, which showed for the first time that many reservoirs can be modeled as part of the standard basin rainfall-runoff and storage processes. In fact, these results provide the first benchmarks for basins with and without reservoirs: zero-*dor*, small-*dor*, and large-*dor* basin subsets had median NSE values of 0.72, 0.79, and 0.60, respectively. Furthermore, the NSE value for even the most challenging large-*dor* basins in the model over the CONUS (0.64) was still comparable to that of the current operational hydrologic model, SAC-SMA, trained and tested only with reference basins (0.65) (Feng et al., 2020a), which further highlights the effectiveness of LSTM as a competitive option for emulating basins with reservoirs for large-scale hydrologic modeling.

Our results provided us with a coherent modeling strategy and some useful lessons. We showed that zero-*dor* and small-*dor* basins behave characteristically differently (and are also different from CAMELS reference basins), which strongly suggests that we cannot simply ignore smaller reservoirs out of convenience and treat them as natural flow, the standard practice in some process-based models. If using a data-driven model, the most beneficial strategy we determined for small reservoirs was to include their reservoir attributes and train a lumped, uniform model that simulated them as part of the basin rainfall-runoff processes. We showed that basins with different *dor* values can be trained together over a large dataset to obtain record-level modeling performance, a strategy which could greatly simplify the modeling process. If using a process-based model, the corresponding approach may be to modify parameters in the model, e.g., linear reservoir parameters, to represent the impacts of smaller reservoirs. The LSTM-based model obtained the best performance in small-*dor* basins without diversion, especially for those with reservoirs for water supply and recreation. For the large-*dor* reservoirs of certain types, i.e., fire protection or farm ponds, hydroelectric, and irrigation dams which are most difficult to model, we may adopt a mixed approach to represent them separately. Considering LSTM is already very strong with respect to feature extraction, it is likely that more relevant information, e.g., electricity prices or irrigation water demand, will be needed to improve their simulation. This paper is the first time such a systematic analysis has been provided from a data-driven perspective.

Our PUB tests advised us of the most important factor in LSTM-based modeling of dammed basins: there must be sufficient representation of small-*dor* and large-*dor* basins in the training set. Dammed and undammed basins behave characteristically differently, and migrating models between them can be dangerous: when a model trained only on CAMELS reference basins or zero-*dor* basins was tested on basins with dams present, we encountered catastrophic failures. We showed that pooling all data together for model training tended to improve results, and even when it did not (likely due to insufficient input information and very heterogeneous training data bringing in noise), the inclusion of training data from other scenarios still did not significantly jeopardize the results.

**Acknowledgements**

The lead author, Wenyu Ouyang, was supported by the China Scholarship Council for one year of study at the Pennsylvania State University. Chaopeng Shen was partially supported by National Science Foundation OAC #1940190. We thank the anonymous reviewers whose comments have helped to substantially improve the manuscript. The list of basins along with their *dor* and NSE values from the LSTM-CONUS model are provided as an attachment in the Appendix. Forcing data used in this study are available from the Daymet dataset website (https://daymet.ornl.gov/); The GAGES-II dataset can be downloaded from the U.S. Geological Survey (USGS) GAGES-II website (https://water.usgs.gov/GIS/metadata/usgswrd/XML/gagesII_Sept2011.xml); Streamflow data can be downloaded from USGS Water Data for the Nation website (http://dx.doi.org/10.5066/F7P55KJN); Reservoir attribution data can be downloaded at National Inventory of Dams website (https://nid.sec.usace.army.mil) from U.S. Army Corps of Engineers; The CAMELS dataset can be downloaded at CAMELS website (http://dx.doi.org/10.5065/D6G73C3Q) from the U.S. National Center for Atmospheric Research. The entire project code is available at GitHub (https://github.com/OuyangWenyu/HydroSPDB). We appreciate the LSTM code at GitHub (https://github.com/mhpi/hydroDL). Many thanks to Google Earth Engine regarding the data processes for the forcing data used in this study.

**Appendix**

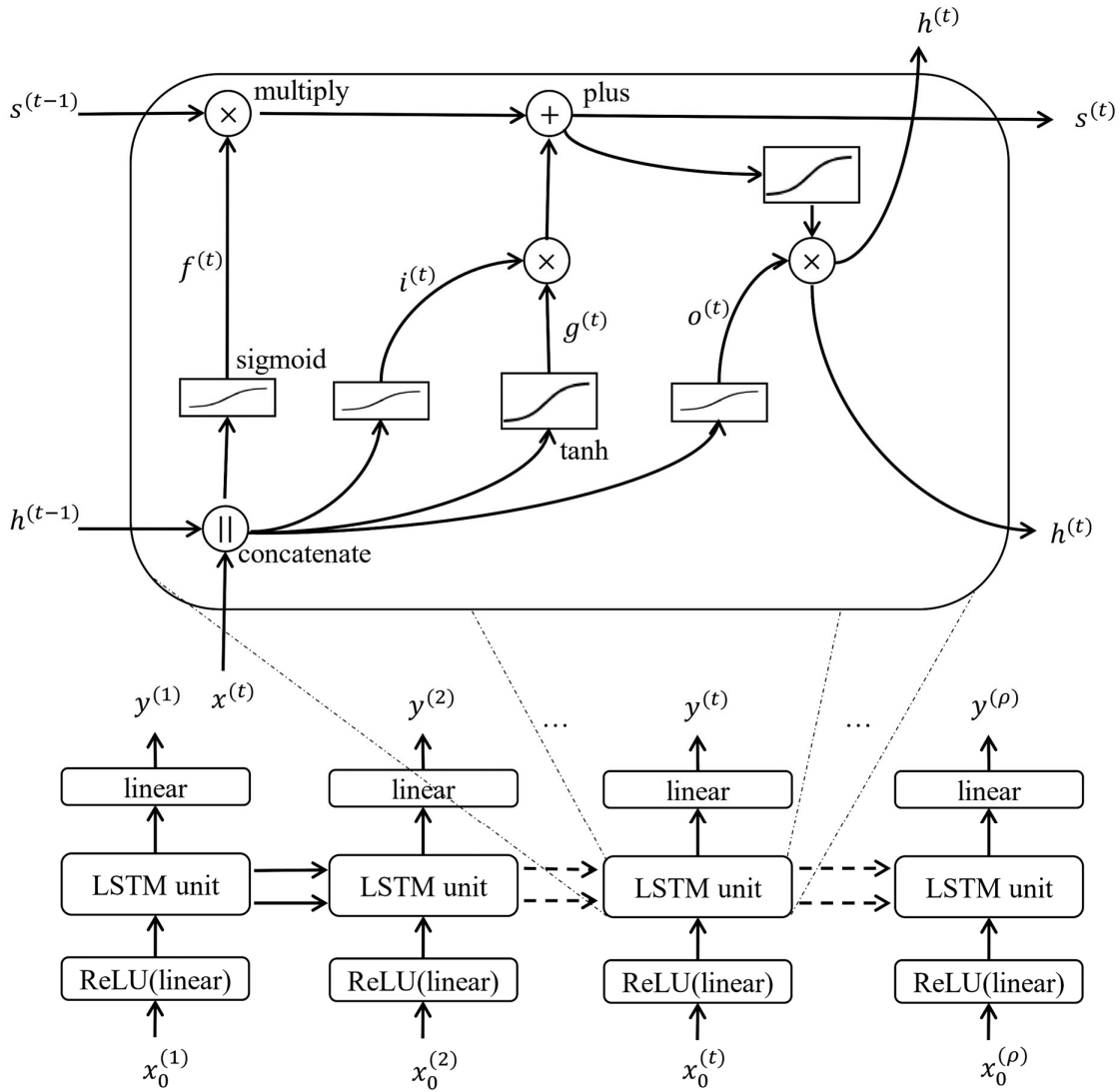

Figure S1. The illustration of the LSTM-based model structure and its unit. $x_0^{(t)}$ is the vector of raw inputs for the time step t, $\rho$ is the length of time sequence of LSTM in the training period. ReLU(linear) is the rectified linear unit, $x^{(t)}$ is the input vector to the LSTM cell, $g^{(t)}$ is the *output of the input node, $f^{(t)}$, $i^{(t)}$, $o^{(t)}$ are the forget, input and output gates, respectively, $s^{(t)}$ represents the states of memory cells, $h^{(t)}$ represents hidden states, and $y^{(t)}$ is the predicted output which is compared to streamflow observations.*

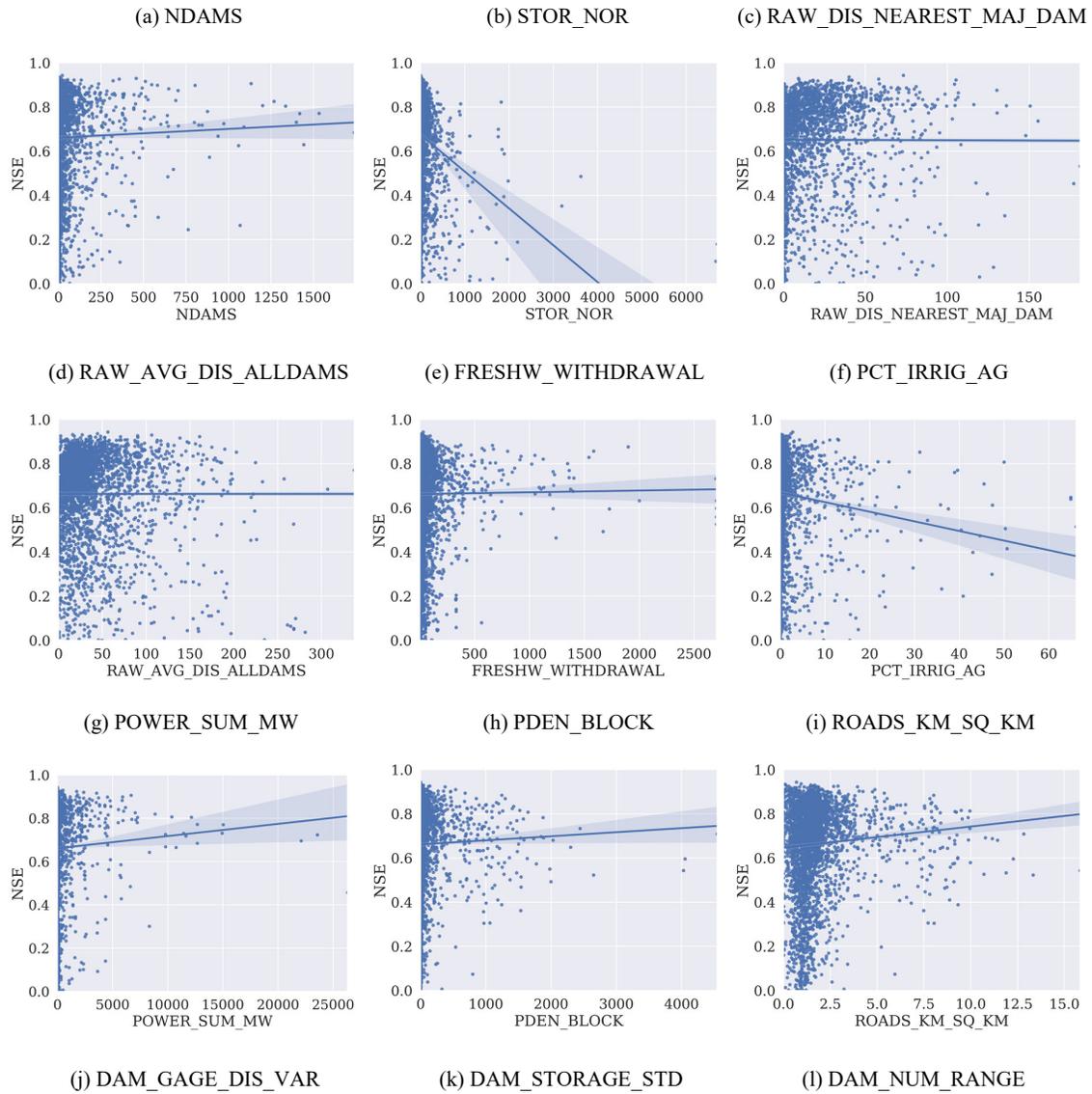

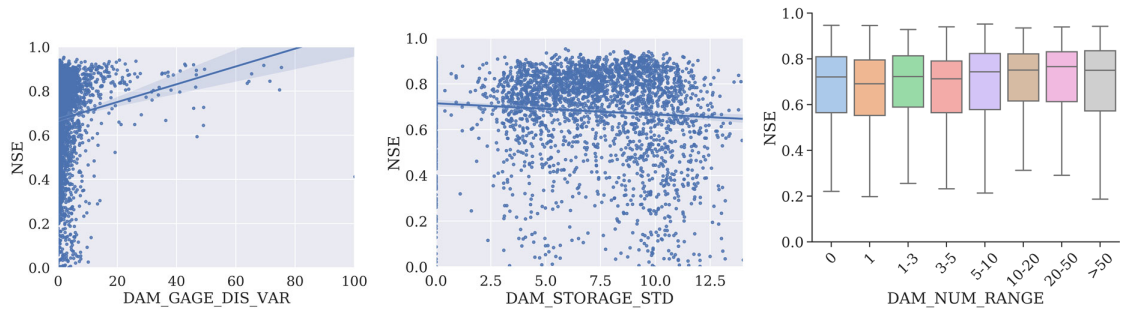

*Figure S2. Scatter plots (subfigure a-k) and a boxplot (subfigure l) for relationships between NSE values (≥0) from the LSTM-CONUS model and some reservoir-related attributes. There are many attributes potentially impacting the performance of the LSTM-based model. We analyzed the information about dams and other anthropogenic hydrologic modifications in the basin in the GAGES-II dataset. (a) NDAMS: number of dams in a basin; (b) STOR_NOR: dam normal storage in a basin, megaliters total storage per sq km; (c) RAW_DIS_NEAREST_MAJ_DAM: raw straight line distance of gage location to nearest major dam in watershed, km. Major dams are defined as being >= 50 feet in height (15m) or having storage >= 5,000 acre feet in GAGES-II; (d) RAW_AVG_DIS_ALLDAMS: raw average straight line distance of gage location to all dams in watershed, km; (e) FRESHW_WITHDRAWAL: freshwater withdrawal, megaliters (1000 cubic meters) per year per sq km; (f) PCT_IRRIG_AG: percent of watershed in irrigated agriculture; (g) POWER_SUM_MW: sum of MW operating capability of electric generation power plants in watershed of type "coal", "gas", "nuclear", "petro", or "water"; (h) PDEN_BLOCK: population density in the watershed, persons per sq km; (i) ROADS_KM_SQ_KM: road density, km of roads per watershed sq km; (j) DAM_GAGE_DIS_VAR: the coefficient of variation of the distances from each dam to the gage location in a basin; (k) DAM_STORAGE_STD: the standard deviation (std) of the normal storages (stor) of reservoirs in a basin; we set std(log(stor+1)) as the x-axis variable; "log" means the natural logarithm; (l) DAM_NUM_RANGE: the ranges of dam numbers -- 0, 1, (1, 3], (3, 5], (5, 10], (10, 20], (20, 50], >50, where "(]" means a left side half open interval; the correspond numbers of basins in each range: 610, 362, 428, 284, 375, 442, 437, 619.*

Table S1. Reservoir simulation results in the literature that do not use recent observations (i.e. data assimilation or data integration). For comparison, our median NSE values reported here were 0.74 for the whole set and 0.78 for basins with small reservoirs. For comparability, we did not include papers that used continual inputs of recent observations of inflow, outflow, or storage.

| Reference | Metric | Description |
| --- | --- | --- |
| Shin et al. (2019) | No NSE reported. Correlations of monthly outflow for the new scheme ($R_{new}$) ranged from -0.07 to 0.63 with a median of 0.25, which were higher than Hanasaki et al. (2006) and Biemans et al. (2011) schemes. | high-resolution continental-scale reservoir scheme (grid-centric) which improved the simulations of reservoirs greatly over the contiguous United States. Tested over six reservoirs in the Missouri, Sacramento, Columbia, San Joaquin, and Colorado River Basins |
| Voison et al. (2013) | Best monthly NSE of regulated flow is 0.62. Negative NSEs for two other locations | An improved grid-centric reservoir formulation to the heavily dammed Columbia River Basin. Authors showed performance metrics for monthly regulated flow at three locations. |
| Wu and Chen (2012) | NSE of outflow ≈ 0.36 | A reservoir operation scheme to decide outflow and its distribution on hydropower, water supply and impoundment purposes according to the inflow and storage. Authors calibrated the coefficients of equations in the new scheme |

| | | during 1965-1984 and validated the scheme in the period 1987-1988 for the Xinfengjiang reservoir |
|---|---|---|
| Kim et al., (2020) | Positive monthly NSEs of daily runoff discharges for real scheduled release; most simulated releases brought negative NSEs (reading off Figure 7) | A grid-centric scheme inside the NWM. Tested on four locations and 21 hydrographs. An NSE of 0.78 was reported for a short period (~11 days) of hourly simulation at one of the locations, but Figure 7 showed mostly negative NSEs. |
| Zajac et al. (2017)$ | Best NSE of streamflow is 0.61 (reading off Figure 6)$ | Global daily streamflow simulations of a spatially distributed LISFLOOD hydrological model in 390 stations during 1980-2013$ |
| Zhao et al. (2016) | NSE of 0.74 and 0.51 for outflow of two reservoirs, respectively.$ | A multi-purpose reservoir module with predefined complex operational rules was integrated into the Distributed Hydrology Soil Vegetation Model (DHSVM). The performance of the model was tested over the upper Brazos River Basin in Texas, where two reservoirs, Lake Whitney and Aquilla Lake, are located$ |
| Payan et al. (2008) | Mean NSE = 0.68 | 46 basins (mostly in France). The quality of the meteorologic dataset in the US, used in this dataset, is potentially lower than the European counterpart. Our work showed that we could obtain NSE=0.84 for CAMELS-GB (Coxon et al., 2020), which has 670 basins from United Kingdom (Ma et al., 2021), while the |
| | | same model with the same training procedure can achieve only 0.74 for CAMELS over CONUS, consistent with other studies. Beck et al., (2020) also showed that NSE for US basins are not higher than global basins. |
| Dang et al. (2020) | A NSE range of 0.68-0.79 for the calibration period, but no value was reported for the validation period | A novel variant of VIC's routing model to simulate the storage dynamics of water reservoirs for the Upper Mekong. However, this study focused on the effect of parameter compensation during calibrating the model or without the reservoir module. Hence, the author did not report on the test period. |

*Table S2. Summary of the forcing and attribute variables used as inputs to the LSTM-based model*

| Variable Type | Variable Name | Description | Unit |
|---|---|---|---|
| Forcing | dayl | Day length | s |
| | prcp | Precipitation | mm/day |
| | srad | Solar radiation | W/m2 |
| | swe | Snow water equivalent | mm |
| | tmax | Maximum temperature | °C |

| | | | | |
|---|---|---|---|---|
| | | tmin | Minimum temperature | ℃ |
| | | vp | Vapor pressure | Pa |
| Attributes | Basic identification and topographic characteristics | DRAIN_SQKM | Watershed drainage area | km² |
| | | ELEV_MEAN_M_BASIN | Mean watershed elevation | m |
| | | SLOPE_PCT | Mean watershed slope | % |
| | | STREAMS_KM_SQ_KM | Stream density | km of streams per watershed km² |
| | Percentages of land cover in the watershed | DEVNLCD06 | Watershed percent "developed" (urban) | - |
| | | FORESTNLCD06 | Watershed percent "forest" | - |
| | | PLANTNLCD06 | Watershed percent "planted/cultivated" (agriculture) | - |
| | | WATERNLCD06 | Watershed percent Open Water | - |
| | | SNOWICENLCD06 | Watershed percent Perennial Ice/Snow | - |
| | | BARRENNLCD06 | Watershed percent Natural Barren | - |
| | | SHRUBNLCD06 | Watershed percent Shrubland | - |
| | | GRASSNLCD06 | Watershed percent Herbaceous (grassland) | - |
| | | WOODYWETNLCD06 | Watershed percent Woody Wetlands | - |
| | | EMERGWETNLCD06 | Watershed percent Emergent Herbaceous Wetlands | - |
| | Soil characteristics | AWCAVE | Average value for the range of available water capacity for the soil layer or horizon | inches of water per inches of |

|  |  |  | soil depth |
|---|---|---|---|
|  | PERMAVE | Average permeability | inches/h |
|  | BDAVE | Average value of bulk density | g/cm$^3$ |
|  | ROCKDEPAVE | Average value of total soil thickness examined | inches |
| Geological characteristics | GEOL_REEDBUSH_DOM | Dominant (highest percent of area) geology | - |
|  | GEOL_REEDBUSH_DOM_PCT | Percentage of the watershed covered by the dominant geology type | - |
| Local and cumulative dam variables | NDAMS_2009 | Number of dams in watershed | - |
|  | STOR_NOR_2009 | Dam storage in watershed ("NORMAL_STORAGE") | megaliters/km$^2$ |
|  | RAW_DIS_NEAREST_MAJ_DAM | Raw straightline distance of gage location to nearest major dam in watershed. | km |
| Other disturbance variables | CANALS_PCT | Percent of stream kilometers coded as "Canal", "Ditch", or "Pipeline" | - |
|  | RAW_DIS_NEAREST_CANAL | Raw straightline distance of gage location to nearest canal/ditch/pipeline in watershed | km |
|  | FRESHW_WITHDRAWAL | Freshwater withdrawal megaliters per year per sqkm | 1000 m$^3$ |
|  | POWER_SUM_MW | Sum of operating capability of electric generation power plants in watershed of type "coal", "gas", "nuclear", "petro", or "water" | MW |
|  | PDEN_2000_BLOCK | Population density in the watershed | persons/km$^2$ |
|  | ROADS_KM_SQ_KM | Road density | km of roads |

| | IMPNLCD06 | Watershed percent impervious surfaces | per watershed km$^2$ % |
|---|---|---|---|

*Table S3. Detailed ensemble results of LSTM-based models in this study*

| Model | Section in the "Experiments" | Random seed | NSE median | Ensemble NSE median |
|---|---|---|---|---|
| LSTM-CONUS | 2.4.1 | 123 | 0.71 | 0.74 |
| | | 1234 | 0.71 | |
| | | 12345 | 0.72 | |
| | | 111 | 0.69 | |
| | | 1111 | 0.72 | |
| | | 11111 | 0.71 | |
| LSTM-CAMELS | 2.4.1 | 123 | 0.73 | 0.75 |
| | | 1234 | 0.74 | |
| | | 12345 | 0.74 | |
| | | 111 | 0.74 | |
| | | 1111 | 0.68 | |
| | | 11111 | 0.73 | |
| LSTM-Z | 2.4.3 | 123 | 0.69 | 0.72 |
| | | 1234 | 0.65 | |
| | | 12345 | 0.71 | |
| | | 111 | 0.69 | |
| | | 1111 | 0.70 | |
| | | 11111 | 0.68 | |
| LSTM-S | 2.4.3 | 123 | 0.77 | 0.79 |
| | | 1234 | 0.77 | |
| | | 12345 | 0.78 | |
| | | 111 | 0.78 | |
| | | 1111 | 0.76 | |
| | | 11111 | 0.76 | |
| LSTM-L | 2.4.3 | 123 | 0.52 | 0.60 |
| | | 1234 | 0.58 | |
| | | 12345 | 0.57 | |
| | | 111 | 0.54 | |
| | | 1111 | 0.59 | |
| | | 11111 | 0.59 | |
| LSTM-ZS | 2.4.3 | 123 | 0.76 | 0.77 |
| | | 1234 | 0.74 | |
| | | 12345 | 0.75 | |
| | | 111 | 0.76 | |
| | | 1111 | 0.77 | |
| | | 11111 | 0.76 | |
| LSTM-ZL | 2.4.3 | 123 | 0.64 | 0.66 |
| | | 1234 | 0.63 | |

| | | 12345 | 0.64 | |
| | | 111 | 0.63 | |
| | | 1111 | 0.64 | |
| | | 11111 | 0.63 | |
| LSTM-SL | 2.4.3 | 123 | 0.72 | 0.75 |
| | | 1234 | 0.73 | |
| | | 12345 | 0.72 | |
| | | 111 | 0.72 | |
| | | 1111 | 0.72 | |
| | | 11111 | 0.72 | |

*Table S4. Ensemble testing results of basins with different dor ranges in different models (Section 3.3)*

| sub-experiment ID | Test basins (number of basins) | Training models | median NSE |
| --- | --- | --- | --- |
| 1 | zero-*dor* basins (610) | LSTM-Z | 0.72 |
| | | LSTM-ZS | 0.72 |
| | | LSTM-ZL | 0.71 |
| | | LSTM-CONUS | 0.72 |
| 2 | small-*dor* basins | LSTM-S | 0.79 |
|   | (1762) | LSTM-ZS | 0.79 |
|   |   | LSTM-SL | 0.78 |
|   |   | LSTM-CONUS | 0.77 |
| 3 | large-*dor* basins (1185) | LSTM-L | 0.60 |
|   |   | LSTM-ZL | 0.63 |
|   |   | LSTM-SL | 0.64 |
|   |   | LSTM-CONUS | 0.64 |

## References

Addor, N., Newman, A.J., Mizukami, N., Clark, M.P., 2017. The CAMELS data set: catchment attributes and meteorology for large-sample studies. Hydrol. Earth Syst. Sci. 21, 5293–5313. https://doi.org/10.5194/hess-21-5293-2017

Ayzel, G., Kurochkina, L., Kazakov, E., Zhuravlev, S., 2020. Streamflow prediction in ungauged basins: benchmarking the efficiency of deep learning, in: E3S Web of Conferences. EDP Sciences, p. 01001. https://doi.org/10.1051/e3sconf/202016301001

Beck, H.E., Pan, M., Lin, P., Seibert, J., van Dijk, A.I.J.M., Wood, E.F., 2020. Global Fully Distributed Parameter Regionalization Based on Observed Streamflow From 4,229 Headwater Catchments. J. Geophys. Res. Atmospheres 125, e2019JD031485. https://doi.org/10.1029/2019JD031485

Biemans, H., Haddeland, I., Kabat, P., Ludwig, F., Hutjes, R.W.A., Heinke, J., von Bloh, W., Gerten, D., 2011. Impact of reservoirs on river discharge and irrigation water supply during the 20th century. Water Resour. Res. 47. https://doi.org/10.1029/2009WR008929

Bindas, T., Shen, C., Bian, Y., 2020. Routing flood waves through the river network utilizing physics-guided machine learning and the Muskingum-Cunge Method, in: American Geophysical Union (AGU). Presented at the AGU Fall Meeting 2020, American Geophysical Union (AGU).

Coxon, G., Addor, N., Bloomfield, J.P., Freer, J., Fry, M., Hannaford, J., Howden, N.J.K., Lane, R., Lewis, M., Robinson, E.L., Wagener, T., Woods, R., 2020. CAMELS-GB: hydrometeorological time series and landscape attributes for 671 catchments in Great Britain. Earth Syst. Sci. Data 12, 2459–2483. https://doi.org/10.5194/essd-12-2459-2020

Dang, T.D., Chowdhury, A.F.M.K., Galelli, S., 2020. On the representation of water reservoir storage and operations in large-scale hydrological models: implications on model parameterization and climate change impact assessments. Hydrol. Earth Syst. Sci. 24, 397–416. https://doi.org/10.5194/hess-24-397-2020

Ehsani, N., Fekete, B.M., Vörösmarty, C.J., Tessler, Z.D., 2016. A neural network based

general reservoir operation scheme. Stoch. Environ. Res. Risk Assess. 30, 1151–1166. https://doi.org/10.1007/s00477-015-1147-9

Falcone, J.A., 2011. GAGES-II: Geospatial Attributes of Gages for Evaluating Streamflow (Report). Reston, VA. https://doi.org/10.3133/70046617

Fang, K., Kifer, D., Lawson, K., Shen, C., 2020. Evaluating the Potential and Challenges of an Uncertainty Quantification Method for Long Short-Term Memory Models for Soil Moisture Predictions. Water Resour. Res. 56, e2020WR028095. https://doi.org/10.1029/2020WR028095

Fang, K., Shen, C., 2020. Near-Real-Time Forecast of Satellite-Based Soil Moisture Using Long Short-Term Memory with an Adaptive Data Integration Kernel. J. Hydrometeorol. 21, 399–413. https://doi.org/10.1175/JHM-D-19-0169.1

Feng, D., Fang, K., Shen, C., 2020a. Enhancing Streamflow Forecast and Extracting Insights Using Long-Short Term Memory Networks With Data Integration at Continental Scales. Water Resour. Res. 56, e2019WR026793. https://doi.org/10.1029/2019WR026793

Feng, D., Lawson, K., Shen, C., 2020b. Prediction in ungauged regions with sparse flow duration curves and input-selection ensemble modeling. ArXiv Prepr. ArXiv201113380.

Gal, Y., Ghahramani, Z., 2016. Dropout as a Bayesian approximation: representing model uncertainty in deep learning, in: Proceedings of the 33rd International Conference on International Conference on Machine Learning - Volume 48, ICML'16. JMLR.org, New York, NY, USA, pp. 1050–1059.

Gauch, M., Kratzert, F., Klotz, D., Nearing, G., Lin, J., Hochreiter, S., 2020. Rainfall–Runoff Prediction at Multiple Timescales with a Single Long Short-Term Memory Network. Hydrol. Earth Syst. Sci. Discuss. 2020, 1–25. https://doi.org/10.5194/hess-2020-540

Giuliani, M., Castelletti, A., Pianosi, F., Mason, E., Reed, P.M., 2016. Curses, tradeoffs, and scalable management: Advancing evolutionary multiobjective direct policy search to improve water reservoir operations. J. Water Resour. Plan. Manag. 142, 04015050. https://doi.org/10.1061/(ASCE)WR.1943-5452.0000570

Giuliani, M., Herman, J.D., Castelletti, A., Reed, P., 2014. Many-objective reservoir policy identification and refinement to reduce policy inertia and myopia in water management. Water Resour. Res. 50, 3355–3377. https://doi.org/10.1002/2013WR014700

Gochis, D., Barlage, M., Dugger, A., FitzGerald, K., Karsten, L., McAllister, M., McCreight, J., Mills, J., RafieeiNasab, A., Read, L., others, 2018. The WRF-Hydro modeling system technical description,(Version 5.0). NCAR Tech. Note 107.

Gorelick, N., Hancher, M., Dixon, M., Ilyushchenko, S., Thau, D., Moore, R., 2017. Google Earth Engine: Planetary-scale geospatial analysis for everyone. Big Remote. Sensed Data Tools Appl. Exp. 202, 18–27. https://doi.org/10.1016/j.rse.2017.06.031

Grill, G., Lehner, B., Thieme, M., Geenen, B., Tickner, D., Antonelli, F., Babu, S., Borrelli, P., Cheng, L., Crochetiere, H., Ehalt Macedo, H., Filgueiras, R., Goichot, M., Higgins, J., Hogan, Z., Lip, B., McClain, M.E., Meng, J., Mulligan, M., Nilsson, C., Olden, J.D., Opperman, J.J., Petry, P., Reidy Liermann, C., Sáenz, L., Salinas-Rodríguez, S., Schelle, P., Schmitt, R.J.P., Snider, J., Tan, F., Tockner, K., Valdujo, P.H., van Soesbergen, A., Zarfl, C., 2019. Mapping the world's free-flowing rivers. Nature 569, 215–221. https://doi.org/10.1038/s41586-019-1111-9

Gupta, H.V., Kling, H., Yilmaz, K.K., Martinez, G.F., 2009. Decomposition of the mean squared error and NSE performance criteria: Implications for improving hydrological modelling. J. Hydrol. 377, 80–91. https://doi.org/10.1016/j.jhydrol.2009.08.003

Gutenson, J.L., Tavakoly, A.A., Wahl, M.D., Follum, M.L., 2020. Comparison of generalized non-data-driven lake and reservoir routing models for global-scale hydrologic forecasting of reservoir outflow at diurnal time steps. Hydrol. Earth Syst. Sci. 24, 2711–2729. https://doi.org/10.5194/hess-24-2711-2020

Hanasaki, N., Kanae, S., Oki, T., 2006. A reservoir operation scheme for global river routing models. J. Hydrol. 327, 22–41. https://doi.org/10.1016/j.jhydrol.2005.11.011

Hochreiter, S., 1998. The Vanishing Gradient Problem During Learning Recurrent Neural Nets and Problem Solutions. Int. J. Uncertain. Fuzziness Knowl.-Based Syst. 06, 107–116. https://doi.org/10.1142/S0218488598000094

Hochreiter, S., Schmidhuber, J., 1997. Long Short-Term Memory. Neural Comput. 9, 1735–1780. https://doi.org/10/bxd65w

International Rivers, 2007. Damming Statistics [WWW Document]. Int. Rivers. URL https://archive.internationalrivers.org/damming-statistics

Kim, J., Read, L., Johnson, L.E., Gochis, D., Cifelli, R., Han, H., 2020. An experiment on reservoir representation schemes to improve hydrologic prediction: coupling the national water model with the HEC-ResSim. Hydrol. Sci. J. 65, 1652–1666. https://doi.org/10.1080/02626667.2020.1757677

Klotz, D., Kratzert, F., Gauch, M., Keefe Sampson, A., Klambauer, G., Hochreiter, S., Nearing, G., 2020. Uncertainty Estimation with Deep Learning for Rainfall-Runoff Modelling. ArXiv E-Prints arXiv:2012.14295.

Kratzert, F., Klotz, D., Herrnegger, M., Sampson, A.K., Hochreiter, S., Nearing, G.S., 2019a. Toward Improved Predictions in Ungauged Basins: Exploiting the Power of Machine Learning. Water Resour. Res. 55, 11344–11354. https://doi.org/10.1029/2019WR026065

Kratzert, F., Klotz, D., Hochreiter, S., Nearing, G.S., 2020. A note on leveraging synergy in multiple meteorological datasets with deep learning for rainfall-runoff modeling. Hydrol. Earth Syst. Sci. 2020, 1–26. https://doi.org/10.5194/hess-2020-221

Kratzert, F., Klotz, D., Shalev, G., Klambauer, G., Hochreiter, S., Nearing, G., 2019b. Towards learning universal, regional, and local hydrological behaviors via machine learning applied to large-sample datasets. Hydrol. Earth Syst. Sci. 23, 5089–5110. https://doi.org/10.5194/hess-23-5089-2019

Lawrence, D.M., Fisher, R.A., Koven, C.D., Oleson, K.W., Swenson, S.C., Bonan, G., Collier, N., Ghimire, B., van Kampenhout, L., Kennedy, D., others, 2019. The Community Land Model version 5: Description of new features, benchmarking, and impact of forcing uncertainty. J. Adv. Model. Earth Syst. 11, 4245–4287. https://doi.org/10.1029/2018MS001583

Lehner, B., Liermann, C.R., Revenga, C., Vörösmarty, C., Fekete, B., Crouzet, P., Döll, P., Endejan, M., Frenken, K., Magome, J., Nilsson, C., Robertson, J.C., Rödel, R., Sindorf, N., Wisser, D., 2011. High-resolution mapping of the world's reservoirs and dams for sustainable river-flow management. Front. Ecol. Environ. 9, 494–502. https://doi.org/10.1890/100125

Ma, K., Feng, D., Lawson, K., Tsai, W.-P., Liang, C., Huang, X., Sharma, A., Shen, C., 2021. Transferring hydrologic data across continents -- leveraging data-rich regions to improve hydrologic prediction in data-sparse regions. Water Resour. Res. n/a, e2020WR028600. https://doi.org/10.1029/2020WR028600

McMahon, T.A., Pegram, G.G.S., Vogel, R.M., Peel, M.C., 2007. Revisiting reservoir storage–yield relationships using a global streamflow database. Adv. Water Resour. 30, 1858–1872. https://doi.org/10.1016/j.advwatres.2007.02.003

McManamay, R.A., 2014. Quantifying and generalizing hydrologic responses to dam regulation using a statistical modeling approach. J. Hydrol. 519, 1278–1296. https://doi.org/10.1016/j.jhydrol.2014.08.053

Mulligan, M., van Soesbergen, A., Sáenz, L., 2020. GOODD, a global dataset of more than 38,000 georeferenced dams. Sci. Data 7, 1–8. https://doi.org/10.1038/s41597-020-0362-5

Nash, J.E., Sutcliffe, J.V., 1970. River flow forecasting through conceptual models part I — A discussion of principles. J. Hydrol. 10, 282–290. https://doi.org/10.1016/0022-1694(70)90255-6

Newman, A.J., Clark, M.P., Sampson, K., Wood, A., Hay, L.E., Bock, A., Viger, R.J., Blodgett, D., Brekke, L., Arnold, J.R., Hopson, T., Duan, Q., 2015. Development of a large-sample watershed-scale hydrometeorological data set for the contiguous USA: data set characteristics and assessment of regional variability in hydrologic model

performance. Hydrol. Earth Syst. Sci. 19, 209–223. https://doi.org/10.5194/hess-19-209-2015

Omernik, J.M., Griffith, G.E., 2014. Ecoregions of the conterminous United States: evolution of a hierarchical spatial framework. Environ. Manage. 54, 1249–1266. https://doi.org/10.1007/s00267-014-0364-1

Paszke, A., Gross, S., Chintala, S., Chanan, G., Yang, E., DeVito, Z., Lin, Z., Desmaison, A., Antiga, L., Lerer, A., 2017. Automatic differentiation in pytorch, in: 31st Conference on Neural Information Processing Systems (NIPS 2017). Long Beach, CA, USA.

Patterson, L.A., Doyle, M.W., 2018. A Nationwide Analysis of U.S. Army Corps of Engineers Reservoir Performance in Meeting Operational Targets. JAWRA J. Am. Water Resour. Assoc. 54, 543–564. https://doi.org/10.1111/1752-1688.12622

Payan, J.-L., Perrin, C., Andréassian, V., Michel, C., 2008. How can man-made water reservoirs be accounted for in a lumped rainfall-runoff model? Water Resour. Res. 44. https://doi.org/10.1029/2007WR005971

Read, J.S., Jia, X., Willard, J., Appling, A.P., Zwart, J.A., Oliver, S.K., Karpatne, A., Hansen, G.J., Hanson, P.C., Watkins, W., others, 2019. Process-guided deep learning predictions of lake water temperature. Water Resour. Res. 55, 9173–9190. https://doi.org/10.1029/2019WR024922

Ryan, J.C., Smith, L.C., Cooley, S.W., Pitcher, L.H., Pavelsky, T.M., 2020. Global Characterization of Inland Water Reservoirs Using ICESat-2 Altimetry and Climate Reanalysis. Geophys. Res. Lett. 47, e2020GL088543. https://doi.org/10.1029/2020GL088543

Sampson, A.K., Hale, E., Lambl, D., 2020. Big Data for Specific Places in Hydrologic Modeling, in: American Geophysical Union (AGU). Presented at the AGU Fall Meeting 2020, American Geophysical Union (AGU).

Sauer, V.B., 2002. Standards for the Analysis and Processing of Surface-Water Data and Information Using Electronic Methods (Report No. 2001–4044), Water-Resources Investigations Report. https://doi.org/10.3133/wri20014044

Shen, C., 2018. A transdisciplinary review of deep learning research and its relevance for water resources scientists. Water Resour. Res. 54, 8558–8593. https://doi.org/10.1029/2018WR022643

Shin, S., Pokhrel, Y., Miguez-Macho, G., 2019. High-Resolution Modeling of Reservoir Release and Storage Dynamics at the Continental Scale. Water Resour. Res. 55, 787–810. https://doi.org/10.1029/2018WR023025

Spangler, K.R., Weinberger, K.R., Wellenius, G.A., 2019. Suitability of gridded climate datasets for use in environmental epidemiology. J. Expo. Sci. Environ. Epidemiol. 29, 777–789. https://doi.org/10.1038/s41370-018-0105-2

Thornton, P.E., Thornton, M.M., Mayer, B.W., Wei, Y., Devarakonda, R., Vose, R.S., Cook, R.B., 2016. Daymet: Daily Surface Weather Data on a 1-km Grid for North America, Version 3. ORNL Distributed Active Archive Center. https://doi.org/10.3334/ORNLDAAC/1328

Turner, S.W.D., Doering, K., Voisin, N., 2020. Data-Driven Reservoir Simulation in a Large-Scale Hydrological and Water Resource Model. Water Resour. Res. 56, e2020WR027902. https://doi.org/10.1029/2020WR027902

US Army Corps of Engineers, 2018. National inventory of dams [WWW Document]. URL https://nid.sec.usace.army.mil/

USGS, 2019. National water information system: Web interface [WWW Document]. U. S. Geol. Surv. URL https://waterdata.usgs.gov/nwis?

Voisin, N., Li, H., Ward, D., Huang, M., Wigmosta, M., Leung, L.R., 2013. On an improved sub-regional water resources management representation for integration into earth system models. Hydrol. Earth Syst. Sci. 17, 3605–3622. https://doi.org/10.5194/hess-17-3605-2013

Wu, Y., Chen, J., 2012. An Operation-Based Scheme for a Multiyear and Multipurpose Reservoir to Enhance Macroscale Hydrologic Models. J. Hydrometeorol. 13, 270–283. https://doi.org/10.1175/JHM-D-10-05028.1

Xiang, Z., Yan, J., Demir, I., 2020. A Rainfall-Runoff Model With LSTM-Based Sequence-to-Sequence Learning. Water Resour. Res. 56, e2019WR025326. https://doi.org/10.1029/2019WR025326

Yang, S., Yang, D., Chen, J., Zhao, B., 2019. Real-time reservoir operation using recurrent neural networks and inflow forecast from a distributed hydrological model. J. Hydrol. 579, 124229. https://doi.org/10.1016/j.jhydrol.2019.124229

Yassin, F., Razavi, S., Elshamy, M., Davison, B., Sapriza-Azuri, G., Wheater, H., 2019. Representation and improved parameterization of reservoir operation in hydrological and land-surface models. Hydrol. Earth Syst. Sci. 23, 3735–3764. https://doi.org/10.5194/hess-23-3735-2019

Yates, D., Sieber, J., Purkey, D., Huber-Lee, A., 2005. WEAP21—A demand-, priority-, and preference-driven water planning model: part 1: model characteristics. Water Int. 30, 487–500. https://doi.org/10.1080/02508060508691893

Yilmaz, K.K., Gupta, H.V., Wagener, T., 2008. A process-based diagnostic approach to model evaluation: Application to the NWS distributed hydrologic model. Water Resour. Res. 44. https://doi.org/10.1029/2007WR006716

Zajac, Z., Revilla-Romero, B., Salamon, P., Burek, P., Hirpa, F.A., Beck, H., 2017. The impact of lake and reservoir parameterization on global streamflow simulation. J. Hydrol. 548, 552–568. https://doi.org/10.1016/j.jhydrol.2017.03.022

Zeiler, M.D., 2012. ADADELTA: An Adaptive Learning Rate Method. CoRR abs/1212.5701.

Zhao, G., Gao, H., Naz, B.S., Kao, S.-C., Voisin, N., 2016. Integrating a reservoir regulation scheme into a spatially distributed hydrological model. Adv. Water Resour. 98, 16–31. https://doi.org/10.1016/j.advwatres.2016.10.014